\documentclass[1p, 11pt, authoryear]{elsarticle}

\usepackage{float}
\usepackage{amsmath}
\usepackage{multirow}
\usepackage{booktabs}
\usepackage{array} 
\usepackage{setspace}
\usepackage{amssymb}
\usepackage[linesnumbered,ruled,algonl,vlined]{algorithm2e}
\usepackage{rotating}
\usepackage{hyperref}
\usepackage{tablefootnote}
\usepackage{txfonts}
\usepackage{verbatim}
\usepackage{xcolor}
\usepackage{paralist}
\usepackage{diagbox}
\usepackage{array, multirow,graphicx}
\usepackage{subcaption}
\usepackage{amsmath}
\usepackage{comment}
\usepackage[belowskip=-15pt,aboveskip=0pt]{caption}
\usepackage{lipsum}
\usepackage{enumitem}
\usepackage[figurename=Fig.]{caption}
\usepackage[inkscapelatex=false]{svg}

\newtheorem{defi}{Definition}

\newproof{proof}{Proof}
\usepackage{soul}
\usepackage{xcolor}

\usepackage{pifont}
\newcommand{\cmark}{\text{\ding{51}}}
\newcommand{\xmark}{\text{\ding{55}}}

\usepackage{color, colortbl}
\definecolor{Gray}{gray}{0.9}
\definecolor{DGray}{gray}{0.81}

\newcolumntype{M}[1]{>{\centering\arraybackslash}m{#1}}

\journal{journal}

\begin{document}

\begin{frontmatter}


\title{Application of Long-Short Term Memory and Convolutional Neural Networks for Real-Time Bridge Scour Prediction}

\author{Tahrima Hashem, Negin Yousefpour{$~^{*}$}}

\address{Department of Infrastructure Engineering, The University of Melbourne, VIC, Australia\\}

\cortext[*]{Corresponding Author: negin.yousefpour@unimelb.edu.au}

\begin{abstract}
Scour around bridge piers is a critical challenge for infrastructures around the world. In the absence of analytical models and due to the complexity of the scour process, it is difficult for current empirical methods to achieve accurate predictions. In this paper, we exploit the power of deep learning algorithms to forecast the scour depth variations around bridge piers based on historical sensor monitoring data, including riverbed elevation, flow elevation, and flow velocity. We investigated the performance of Long Short-Term Memory (LSTM) and Convolutional Neural Network (CNN) models for real-time scour forecasting using data collected from bridges in Alaska and Oregon from 2006 to 2021. The LSTM models achieved mean absolute error (MAE) ranging from 0.1m to 0.5m for predicting bed level variations a week in advance, showing a reasonable performance. The Fully Convolutional Network (FCN) variant of CNN outperformed other CNN configurations, showing a comparable performance to LSTMs with significantly lower computational costs. We explored various innovative random-search heuristics for hyperparameter tuning and model optimisation which resulted in reduced computational cost compared to grid-search method. The impact of different combinations of sensor features on scour prediction showed the significance of the historical time series of scour for predicting upcoming events. Overall, this study provides a greater understanding of the potential of Deep Learning algorithms for real-time scour prediction and early warning for bridges with distinct geology, geomorphology and flow characteristics. 
\end{abstract}


\end{frontmatter}

\section{Introduction}\label{sec:intro}
Bridge scour has been a major challenge for transportation systems around the world due to the significant damage and operational disruptions it can cause, and more importantly the risk to public safety. In the US, for example, scour accounts for a significant number of bridge collapses, with more than 260,000 brides listed as vulnerable to scour failure \cite{Yousefpour&Medina}. 
It is difficult to reliably predict the maximum scour depth for a bridge pier due to the complexity of the underlying causes of soil erosion in interaction with flow and structure. Uncertainties in riverbed material, flow, and geomorphological conditions, as well as climate change impacts, are the main challenges in the assessment of scour risk. For the last couple of decades, several research studies ~\cite{Empirical_AP1,Empirical_AP2,liang2019local,pizarro2020science} have focused on developing empirical models to estimate scour depth based on laboratory experiments and field observations. Most of these models often overestimate the depth of the scour (and in some cases underestimate) due to insufficient generalization to various conditions of the riverbed, flow, and structure ~\cite{Survey}. 

Artificial Intelligence (AI) and Machine Learning (ML) for scour depth estimation have been investigated by a range of algorithmic strategies, such as evolutionary computing, fuzzy logic, artificial neural networks, support vector machines, and decision trees (see section ~\ref {sec:lit-review}). These proposed techniques have shown promising results in estimating maximum scour depth, often outperforming traditional empirical equations. However, the performance of these models is still limited by the availability and quality of training data, which are often scarce and may not cover a wide range of geological, hydraulic and geomorphological conditions. Furthermore, these models are typically trained to predict a maximum scour depth for a given flow discharge and cannot be used for dynamic scour prediction and real-time forecasting.

\begin{figure}[t]
    \centering
    \includegraphics[width=0.9\textwidth]{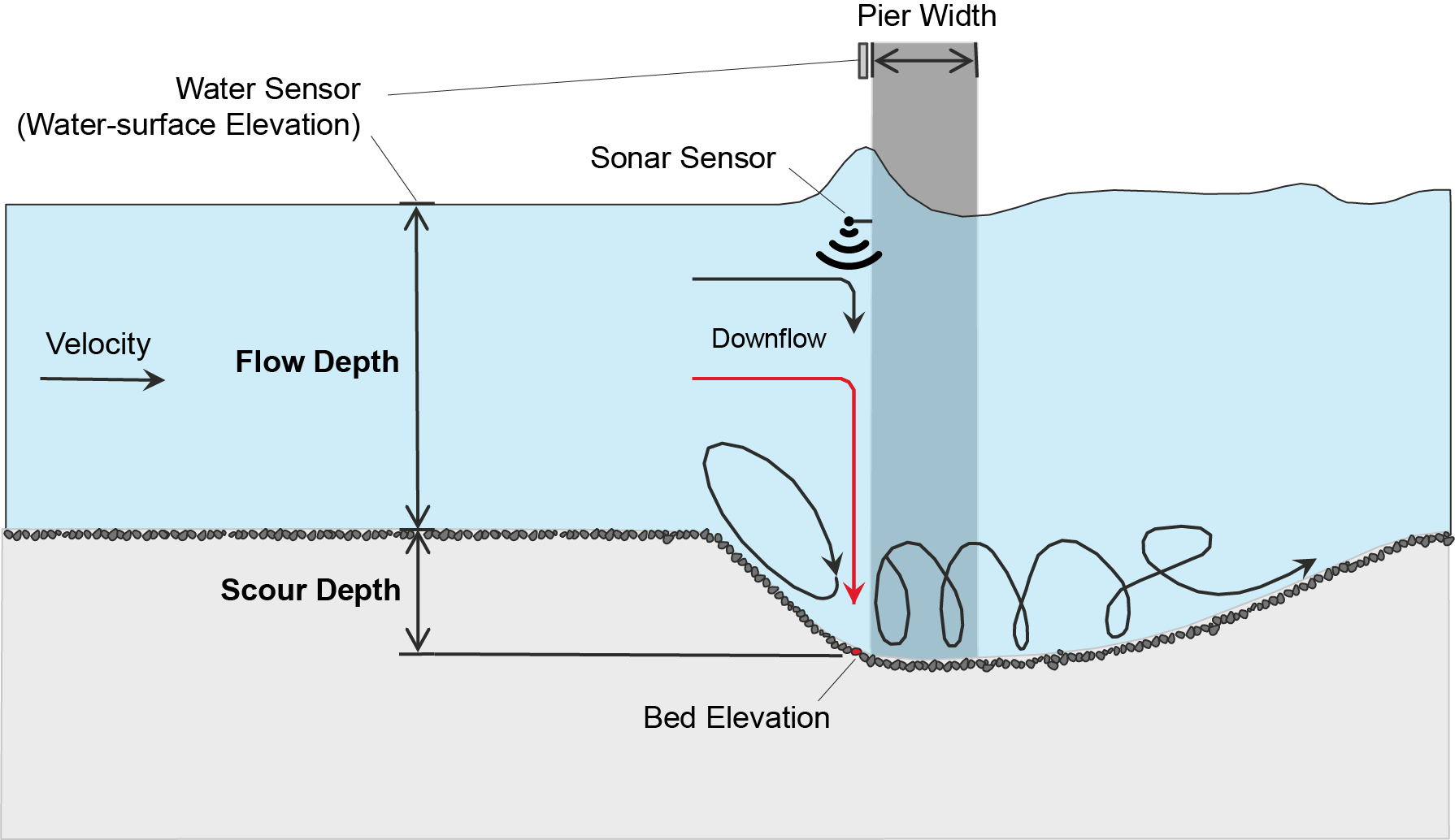}
    \caption{An example of the bridge scour, adopted from \cite{arneson2013evaluating}.}
    \label{fig:1}
\end{figure}

Yousefpour et al. \cite{phase1, phase2} pioneered the application of deep learning (DL) solutions, using Long Short Term Memory (LSTM) networks \cite{hochreiter1997long} for real-time scour forecasting. They leveraged the LSTM's superiority in temporal pattern recognition and capturing the underlying physics without direct feature extraction. They used historical scour monitoring data in Alaska to train DL models and predict future bed elevation variations and upcoming scour events. Fig. \ref{fig:1} shows the process of local scour around a bridge pier; scour happens when sediments are eroded and washed away by strong vortices in interaction with the bridge substructure, often accelerating during floods. Stage (flow/water level) and sonar (bed elevation) sensors are commonly used sensors installed on the bridge piers to monitor scour by collecting real-time flow and bed elevation data. The variants of LSTM models developed in \cite{phase2} showed promising performance in forecasting the scour depth a week in advance, for case-study bridges in Alaska. However, there remain research questions about the capability of these algorithms to be upscaled for implementation in practice and deployment in different locations with variable meteorological, geological, geomorphological, and flow conditions.

In this study, we addressed fundamental research questions about the efficiency and generalization of the LSTM scour forecast models in diverse spatio-temporal domains. The LSTM models are compared against temporal CNN as an alternative deep learning solution, in terms of performance and computational cost. Random-Search (RS) heuristics for hyperparameter (HP) tuning are evaluated ~\cite {RS1} to explore alternate computationally efficient methods to Grid-Search (GS) used in ~\cite{phase2} for model optimisation. We also investigated the impacts of monitored features (input parameters), including stage, sonar, and velocity (discharge), on the performance of scour forecast models. 

To summarise, four research questions were addressed in this study:

(RQ1) \textit{How do LSTM scour forecast models perform in predicting recent years of scour (Alaska) and how well do these models perform in a totally different location (Oregon)?}

(RQ2) \textit{How do other competitor models, including CNN with vanilla and dilated convolutions, compare to LSTMs for the prediction of scour?} 

(RQ3) \textit{What computationally efficient heuristics can be adopted to find optimal model configurations to scour forecast DL models without sacrificing performance?}

(RQ4) \textit{To what extent combinations of different sensor features can impact the performance of scour forecast DL models, and what are the most optimum feature combinations?}

We carefully designed experiments and developed heuristics to investigate the aforementioned research questions as discussed in the following sections. The DL models were trained on the University of Melbourne's Spartan high-performance computing (GPU and CPU) cluster. The models were developed in Python using Tensorflow library~\cite{tensorflow2015-whitepaper}.

\section{Background}\label{sec:lit-review}
\subsection{Maximum Scour Prediction}
 
Since the 1940s, researchers have been using laboratory experiments and real-world observations to formulate empirical equations to estimate maximum scour depth. The famous equations HEC-18 \cite{arneson2013evaluating}, FDOT \cite{FDOTManual2021}, and CIRIA \cite{CIRIA2015} equations were derived by multivariate analysis, considering various factors that impact the scour process. These equations have a large number of input and model parameters and measurement of the inputs is complex. However, often small databases (a few dozen to a few hundred samples) are employed for model calibration \cite{sheppard2011scour}. Also, these empirical equations follow particular mathematical forms, such as linear, exponential, and hyperbolic terms, limiting their flexibility in capturing scour as a nonlinear phenomenon. As a result, the recommended equations often underperform in application to a broad range of scour conditions.

With the surge of data-driven methodologies, more sophisticated approaches from machine learning have been implemented for scour depth estimation, including Support Vector Machines (SVM) \cite{goel2009application,sharafi2016design,najafzadeh2016scour}, Genetic Algorithms \cite{azamathulla2010genetic,Khan2012}, Decision Trees \cite{pal2013pier,tien2020hybrid}, Artificial Neural Networks (ANNs) \cite{choi2006prediction,lee2007neural}, and CNNs \cite{zhang2020prediction}. The results show that machine learning-based methods are usually more accurate than traditional equations in maximum scour depth prediction. However, these ML models face limitations when encountering unseen cases outside the convex hull of the training data,  which can lead to poor generalization. 

\subsection{Real-time Scour Monitoring and Forecast}
Although the worst-case scenario (usually a 1-in-100-year flood) is considered for estimating maximum scour depth in bridge design, many uncertainty factors could still lead to scour-related bridge failures: Larger flood events can occur due to extreme weather conditions and climate change; complexities in the scour process, particularly river sediment material interaction with flow which can result in larger scour depth;  older bridges could fall behind the best recommended practices for scour estimation \cite{khandel2019integrated, lamb2017vulnerability}.  Considering all these factors, the transport authorities in the past two decades have been resorting to regular monitoring to manage scour, including regular inspections and surveys as well as advanced remote sensing.

In recent years, real-time scour monitoring sensors have become a prominent solution installed on many large-scale and critical bridges \cite{lagasse1997instrumentation, briaud2011realtime, lebbe2014failure,PRENDERGAST:2014}. With the development of monitoring approaches, the sensors can facilitate data collection in real-time concerning bed elevation and other flow-related information. Compared to traditional databases \cite{sheppard2011scour}, the sensor data collected is richer and more granular. Therefore, the monitoring data can also be utilized for scour depth prediction and providing early-warning information.

Yousefpour et al. \cite{phase1, phase2} pioneered the idea of incorporating real-time scour monitoring data to develop real-time scour forecast models using LSTMs. They incorporated monitoring data in Alaska (2006 to 2017) collected by the US Geological Survey (USGS) in partnership with the Departments of Transport (DOTs) in the United States. Specifically, historical time series of flow depth and bed elevation were used to train the LSTM models to predict future bed elevation and upcoming scour depth. The proposed model showed promising results in forecasting bed-elevation changes for at least seven days in advance. Compared to traditional empirical equations, these time series forecasting models yielded more reliable predictions of the upcoming scour trends and maximum depth of scour.

Chang et al. \cite{chang2014pier} presented a real-time monitoring system, using micro-cameras mounted on bridge piers, in conjunction with image processing and pattern recognition methodologies, to monitor variations in bed levels. This approach was validated through rigorous laboratory experiments, illustrating that the image data can be transformed to capture the progressive alterations in scour depth. Typically, monitoring sensors are installed on bridges deemed high risk. 

To leverage limited data to better estimate scour risks across a bridge network, Maroni et al. \cite{maroni2021using} propose a scour hazard model tailored for road and railway bridges. By using a Bayesian network, the model decreases uncertainty in scour depth assessments at unmonitored bridges. This approach was demonstrated in a case study involving several Scottish road bridges.

Lin et al. \cite{CNNImageData} developed an early scour warning system installed at Da-Chia Bridge in Taiwan. They used R-CNN (a residual CNN) to read the real-time water level through CCTV images, and dedicated vibration-based arrayed sensors to obtain real-time local bed elevation of the target pier. These two data sources were incorporated to simulate upcoming hydrodynamic flows in fluvial rivers with hydraulic structures (e.g., bridge piers). The general scour is derived from a numerical model, and the local scour is estimated by empirical equations.

\section{Scour Monitoring Data}

\begin{figure}[h]
\centering
\includegraphics[width=1.0\linewidth]{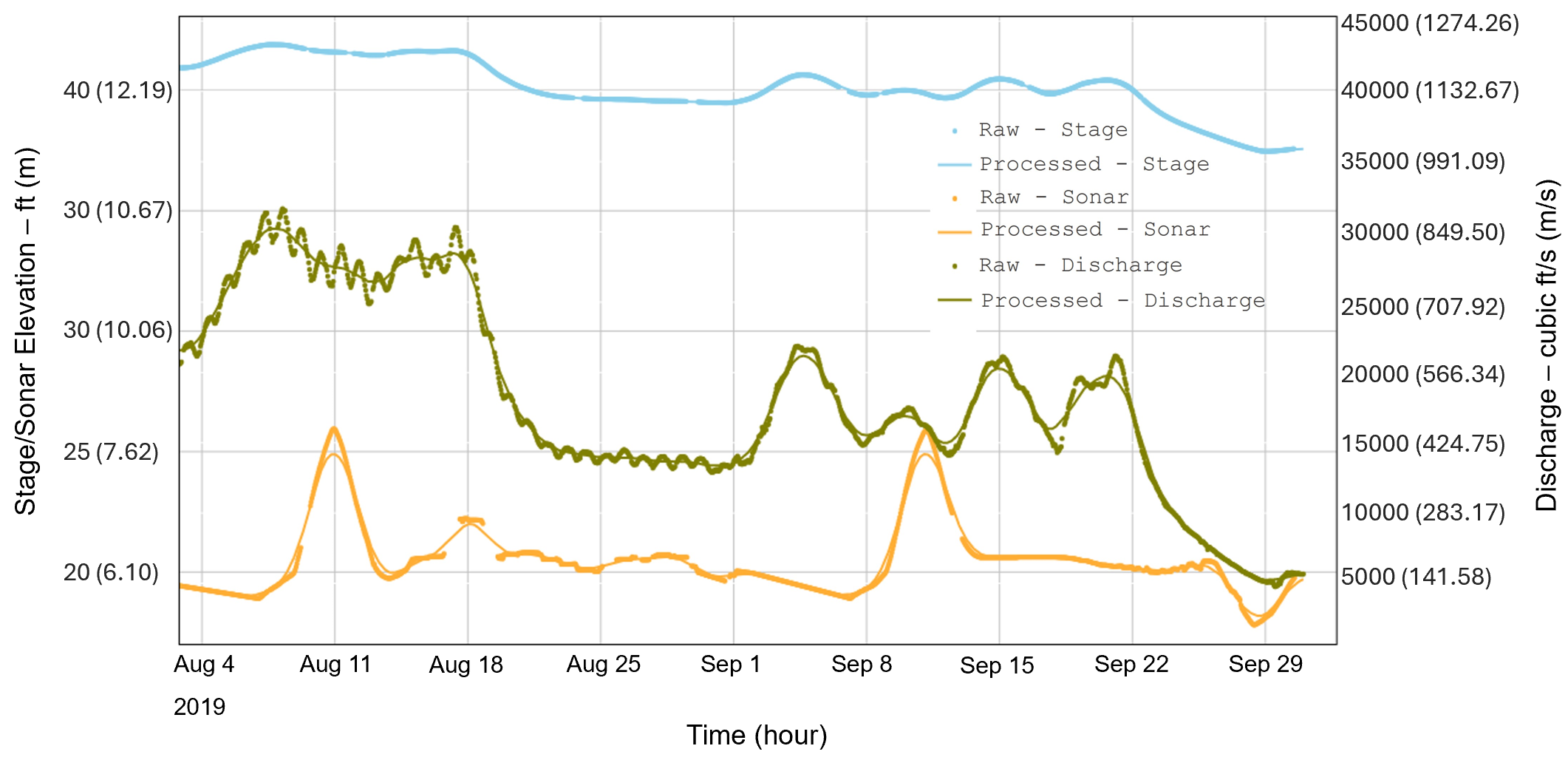}
\caption{Raw vs. Processed data - Alaska $539$ bridge.}
\label{Fig:539_dataexa}
\end{figure}

\begin{figure}[h]
\centering
\includegraphics[width=0.9\linewidth]{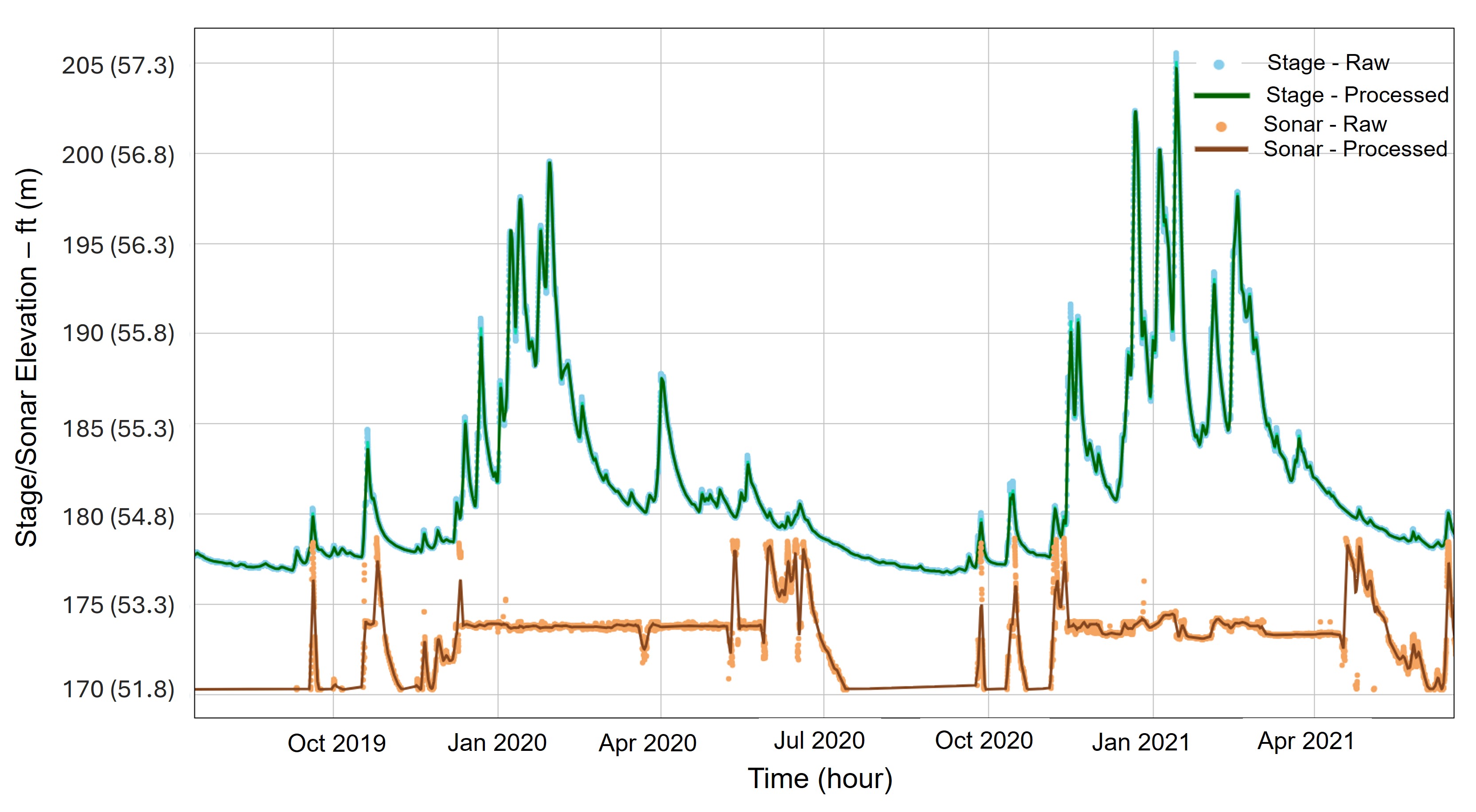}
\caption{Raw vs. Processed data - Oregon Luckiamute Bridge.}
\label{Fig:Luck_dataexa}
\end{figure}

We considered two scour monitoring databases for Alaska and Oregon provided by USGS to develop real-time scour forecast models:
\begin{itemize}
    \item Alaska dataset: This includes previously collected sensor readings data from 2006 to 2017 from bridges over Sheridan River, Knik River, and Chilkat River with IDs $230$, $539$, and $742$, respectively. Readers are referred to \cite{phase2} for full data description. In addition, in this study we incorporated new data collected for $2018$ to $2021$ (for $742$ bridge, $2019$ and $2020$ data are not available). An example of processed versus raw data for Alaska $539$ bridge is provided in Fig. ~\ref{Fig:539_dataexa}.
    \item Oregon dataset: This includes sensor readings from 2018 to 2021 for two bridges over the Trask and Luckiamute rivers (see Fig. ~\ref{Fig:Luck_dataexa}). 
\end{itemize}
For both datasets, each observation corresponds to an hourly recorded reading of the available sensors. We followed the same methodology as \cite{phase1} to denoise, synchronize, and filter the raw monitoring data provided by USGS. 
\vspace{1em}

A strong negative correlation between stage and sonar can be observed for Alaska bridges. This is expected as higher flow velocity and depth result in increased scour depth, for most riverbed sedimentation.
In addition, Alaskan bridges show consistent seasonal scour and filling patterns throughout a year, which is a key characteristic of Live-Bed scouring; each year, major scour followed by filling episodes are observed due to high flows (floods) happening around July/August and September/October.
However, the scour patterns for the two bridges in Oregon are fundamentally different. Trask is a tidal river where bed material follows the tide cycles: scouring during outgoing tides and filling occurs during high tides. Also, the Luckiamute bridge shows a positive correlation trend between stage and sonar, contrary to expectations. Oregon USGS relates this to the presence of bed forms, which are structures such as sand dunes formed by the movement of bed material following the flow direction. 

\vspace{1em}

\section{Approach}

\subsection{Problem Definition}
Given the historical sensor readings, i.e., water level: \textit{Stage}, riverbed elevation: \textit{Sonar}, and flow discharge: \textit{Discharge} over time, we define the scour forecast task as a multivariate time-series prediction problem. 
Following is the mathematical definition of the scour forecast task.

\begin{defi}
Assume a series of past values $\{x_{t-w_{in}+1}, x_{t-w_{in}+2}, \dots, x_t\}$ for $N$ sensors (features) with the historical window $w_{in}$ where $x_i \in \mathbb{R}^N$ and $1 < w_{in} \leq t$. The scour forecasting task is to predict a series of future values, \{$x_{t+1}, x_{t+2}, \dots, x_{t+w_{out}}$\} for the \textit{Sonar} with the forecast window $w_{out}$ where $w_{out} \leq t$.
\end{defi}

\subsection{Feature Engineering}\label{sec:feat_engineering}
In this study, we explored adding Discharge ($X_\textit{discharge}$) in addition to Stage ($X_\textit{stage}$) and Sonar ($X_\textit{sonar}$), incorporated in our previous study \cite{phase2} for Alaskan bridges \footnote{for Oregon bridges, this sensor data was not available}. Discharge represents the water volume passing the river cross section per unit of time. $X_\textit{discharge}$ which is obtained from the mean flow velocity measured by velocimeter sensors multiplied by the river channel cross-section area measured at the bridge location ~\cite{DischargeMA,turnipseed2010discharge}.

We also engineered another feature, the equivalent velocity $X_\textit{eVelocity}$, which is directly proportional to the flow velocity ($V$) following Equations. \ref{eq:gage} to ~\ref{eq:dis_v}. By assuming a constant channel equivalent cross-sectional width $\alpha$, the channel cross section area can be estimated by $\alpha$ x $d$, where $d$ is the flow depth at a bridge pier equal to the difference between Sonar and Stage elevations (Equation ~\ref{eq:gage}).

\begin{align}
    d &= X_\textit{sonar}- X_\textit{stage} \label{eq:gage} \\
    X_\textit{discharge} &= \alpha d V \label{eq:dis}\\
    X_\textit{eVelocity} &=  \alpha V \\
    X_\textit{eVelocity} &= X_\textit{discharge}/(X_\textit{sonar} - X_\textit{stage})\label{eq:dis_v}
\end{align}

The previous study showed that time features did not improve the performance of the LSTM models for Alaskan bridges. However, to enrich the periodicity information in Oregon's scour data, \textit{time} features, namely $\textit{year-sin}$ and $\textit{year-cos}$ are incorporated in this study. These features are calculated by applying \textit{sin}($\cdot$) and \textit{cos}($\cdot$) transformations to the recorded timestep using Equation ~\ref{eq:time_sin} and Equation ~\ref{eq:time_cos}, respectively.


\begin{align}
\textit{year-sin} &= \sin(\textit{TS}.2\pi / \textit{year}) \label{eq:time_sin} \\
\textit{year-cos} &= \cos(\textit{TS}.2\pi / \textit{year}) \label{eq:time_cos}
\end{align}

The features \textit{Sonar}, \textit{Stage}, \textit{Discharge} and \textit{eVelocity} are denoted as \texttt{sN}, \texttt{sT}, \texttt{dC} and \texttt{dV}, respectively.  Time features are denoted as \textit{y}. The impact of various features and their combinations are discussed in Section \ref{sec:wrapper}.

\subsection{Deep Learning For Scour Forecast}\label{sec:DL_models}

\subsubsection{LSTM Models}\label{sec:LSTMmodel}
\begin{figure}[h]
    \centering
    \begin{subfigure}[t]{0.4\textwidth}
        \centering
        \includegraphics[width=1\textwidth]{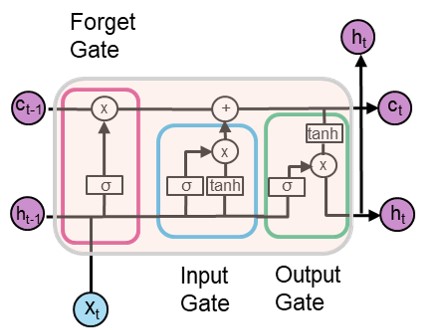}
        \subcaption{}
    \end{subfigure}\\
    \vspace{15pt}
    \begin{subfigure}[t]{0.7\textwidth}
        \centering
        \includegraphics[width=1\textwidth]{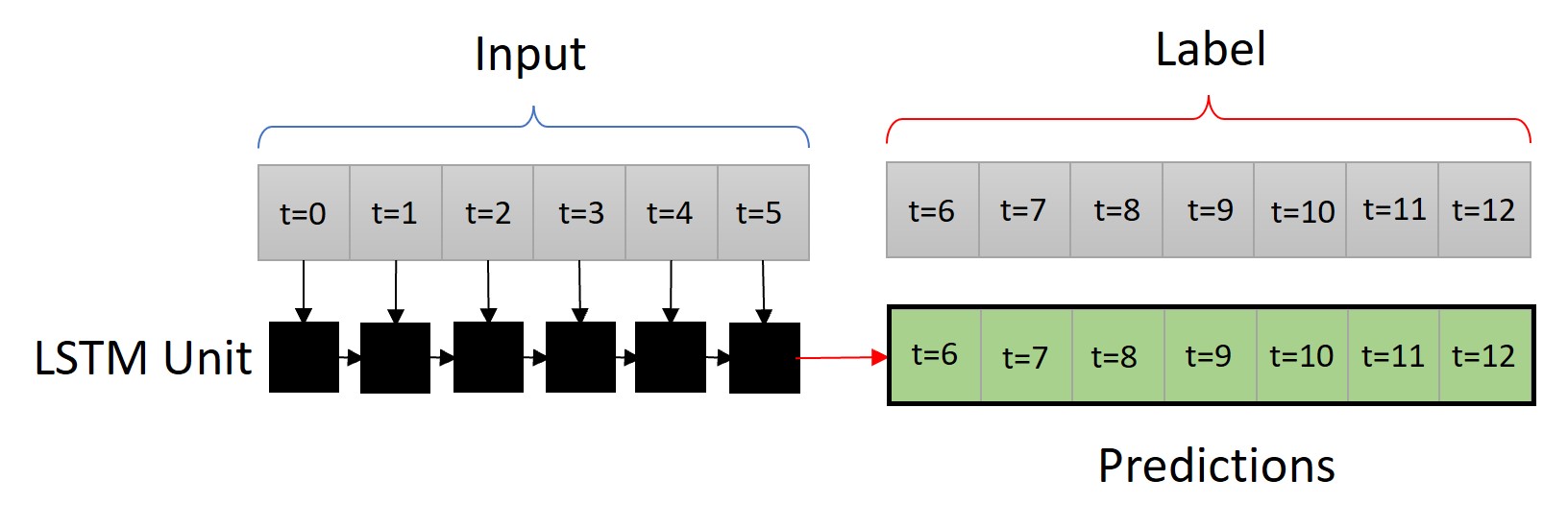}
        \subcaption{}
    \end{subfigure}\\
      \vspace{15pt}
     \begin{subfigure}[t]{0.7\textwidth}
        \centering
        \includegraphics[width=1\textwidth]{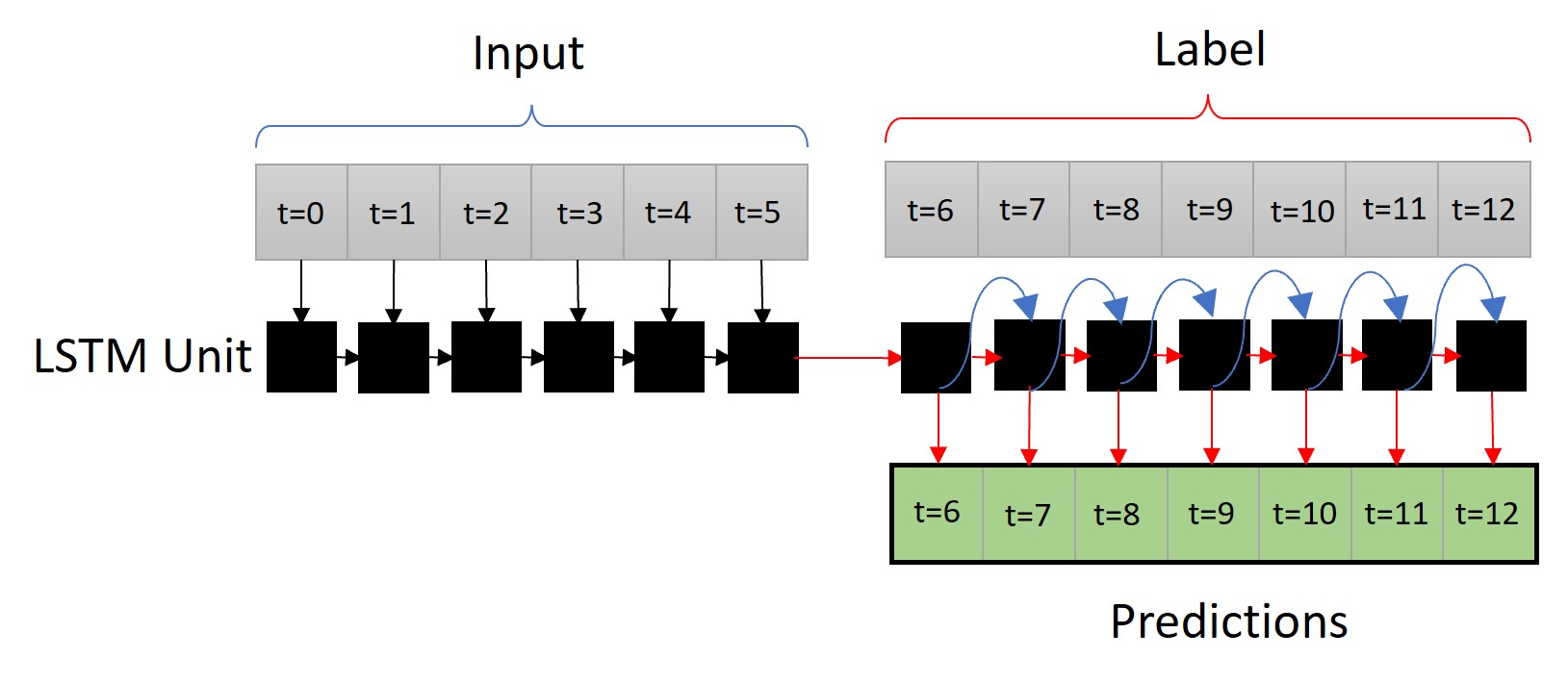}
        \subcaption{}
     \end{subfigure}
       \vspace{15pt}
\caption{(a) LSTM memory unit, (b) Single-Shot and (c) Feedback variants}
\vspace{10pt}
\label{fig:lstmprocess}
\end{figure}

Long Short Term Memory~\cite{hochreiter1997long} is a specialised recurrent neural network (RNN) that is capable of modelling very large input and output sequences and does not suffer from the \textit{gradient vanishing problem} - a typical problem in RNNs~\cite{LSTM1, LSTM2}. 
The principle analogy of LSTM is to remember the relevant short and long-standing information and discard the irrelevant information. Three different gates, i.e., \textit{Forget} gate ($\Gamma^{f}$), \textit{Input} gate ($\Gamma^{i}$) and \textit{Output} gate ($\Gamma^{o}$), enable the LSTM model to achieve this goal as shown in Fig. ~\ref{fig:lstmprocess} (a). In these gates, the degree of relevance of information is quantified using a sigmoid ($\sigma$) function with a value between $0$ and $1$. The larger the value, the higher the relevance. Equations (\ref{EQ:LSTM-1} - \ref{EQ:LSTM-6}) show how the hidden unit output ($a_t$) and LSTM memory cell output ($c_t$) are recurrently computed using these gates. $W$ and $b$ represent the weight and bias matrices.

\begin{align}
\Gamma^{f} &= \sigma(W_f[a_{t-1},x_t]+b_f)
\label{EQ:LSTM-1} \\
\Gamma^{i} &= \sigma(W_i[a_{t-1},x_t]+b_i)
\label{EQ:LSTM-2} \\
\Gamma^{o} &= \sigma(W_o[a_{t-1},x_t]+b_o)
\label{EQ:LSTM-3} \\
\tilde{c_t} &= \tanh(W_c[a_{t-1},x_t]+b_c)
\label{EQ:LSTM-4} \\
c_t &=  \Gamma^{i}\tilde{c_t} + \Gamma^{f}c_{t-1}
\label{EQ:LSTM-5} \\
a_t &= \Gamma^{o}\tanh(c_t) 
\label{EQ:LSTM-6}
\end{align}

\begin{table}[ht]
\centering
\begin{tabular}{c}
      \resizebox{150mm}{!}{\includegraphics{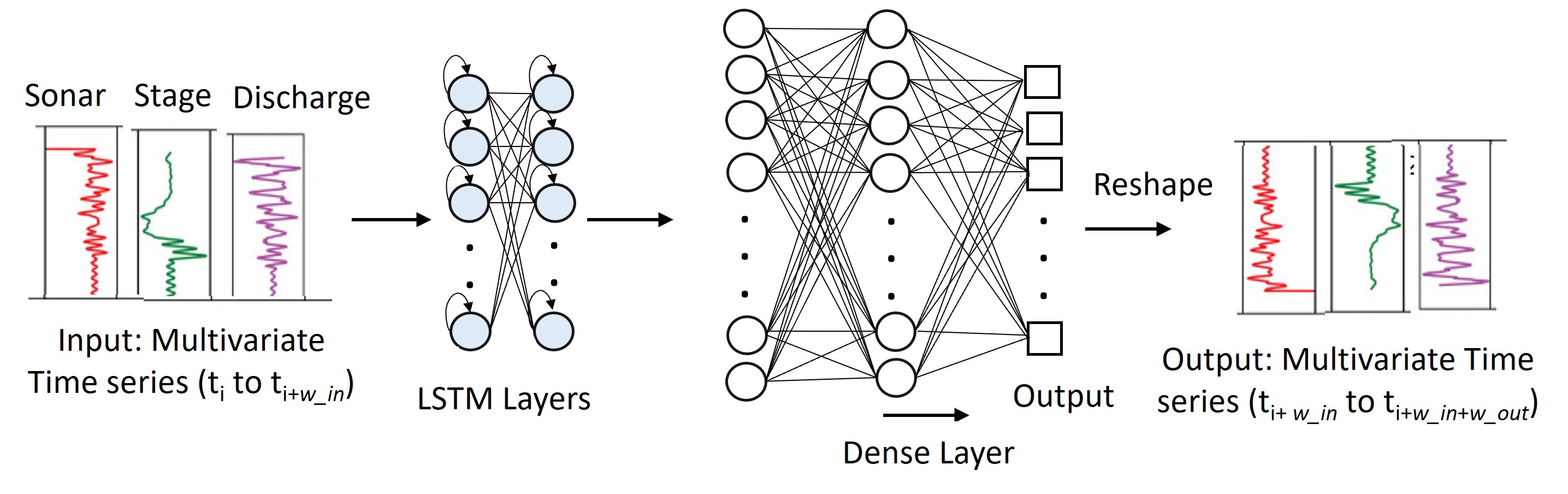}}
\end{tabular}
\captionof{figure}{Multivariate time series modelling using LSTM model}
\label{Fig:LSTM}
\end{table}
\vspace{1em}

Fig. ~\ref{Fig:LSTM} shows the architecture of the LSTM models for the multivariate time series forecasting of scour. The following three variants introduced in \cite{phase2} are also implemented in this study:

\begin{enumerate}
\item Single-Shot (ss): In this model, one LSTM layer has been followed by one Dense and one Reshape layer. This model outputs the predictions for all the timesteps over the forecast window at once, as shown in Fig.~\ref{fig:lstmprocess}(b). 
\item Two-Layer, Single-Shot (ss2): Two LSTM layers are stacked together in this variant. The rest of the setting remains the same as Single-Shot.
\item Feedback (fb): In this variation, we hypothesize that the model's prediction accuracy would get improved, if we input the predicted value (as opposed to the actual value) of each timestep for the prediction of the next timestep. Therefore, in this variant, LSTM predicts one timestep ahead across the entire forecast window. Fig.~\ref{fig:lstmprocess} (c) illustrates the feedback process. The feedback variant also has one LSTM layer, similar to the Single-Shot model.
\end{enumerate}

Model configuration nomenclature for an LSTM model is encoded as follows:

\begin{equation}\label{eq:encoding_config}
    [model]-[w_{in},w_{out}]-[units]-[dropout]
    \nonumber
\end{equation}
where, \textit{model} is the LSTM variant, \textit{$w_{in}$} and \textit{$w_{out}$} are the length of model inputs and outputs, \textit{units} is the number of LSTM units, and \textit{dropout} denotes the dropout rate applied to the model.

\subsubsection{CNN Models}\label{sec:CNN_models}

CNN models proposed in ~\cite{earlyCNN} are widely used in computer vision for tasks like image classification ~\cite{CNN_AlexNet} and object detection ~\cite{CNNobjectdetection}. They offer excellent performance with lower computational requirements. CNNs also excel in time series modelling tasks ~\cite{CNN_timeseries, CNNintimeseries, TCN}. We investigated three CNN variants for multi-variate time-series scour forecasting with an architecture as shown in Fig.~\ref{Fig:CNN}).

\begin{table}[htbp]
\centering
\begin{tabular}{c}
      \resizebox{150mm}{!}{\includegraphics{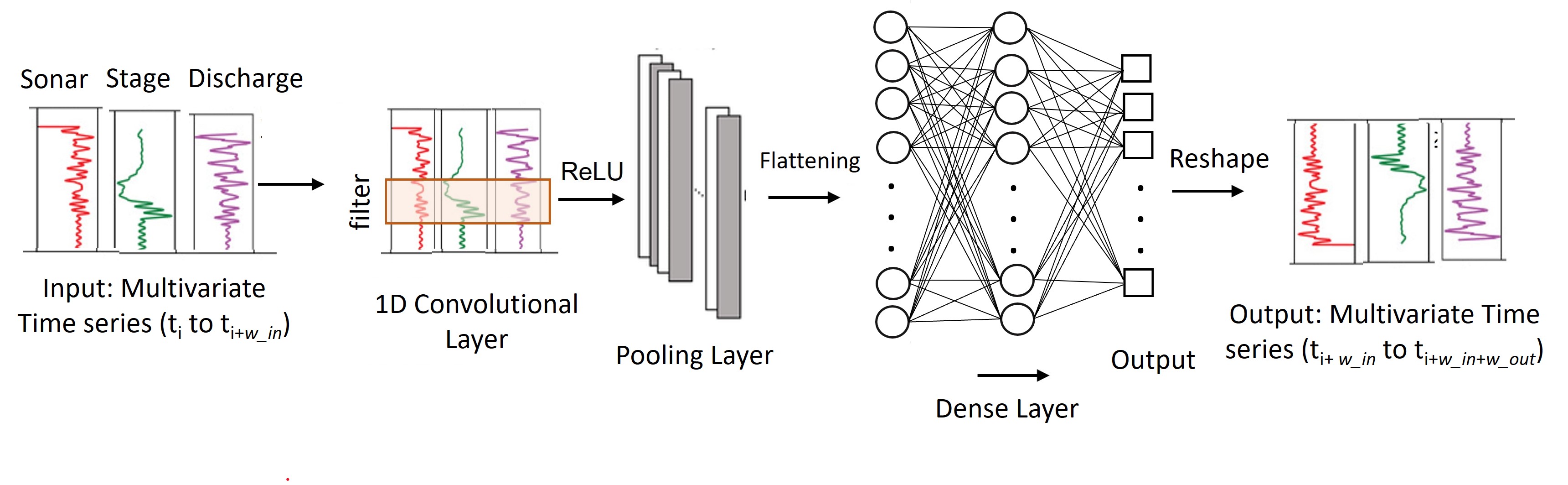}}
\end{tabular}
\captionof{figure}{Multivariate time series modelling using CNN model}
\label{Fig:CNN}
\end{table}
\vspace{1em}

\begin{figure}[htbp]
\centering
\begin{tabular}{cc}
\hspace{-1cm}
      \includegraphics[height=0.22\textwidth]{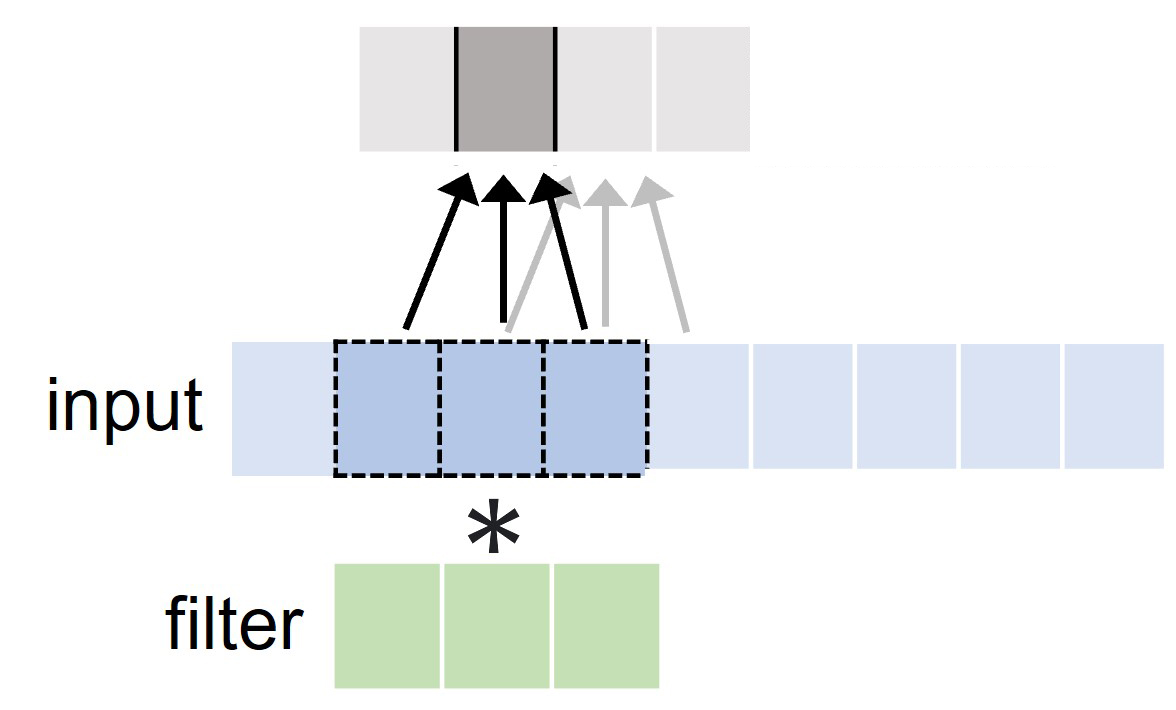} &
      \includegraphics[height=0.22\textwidth]{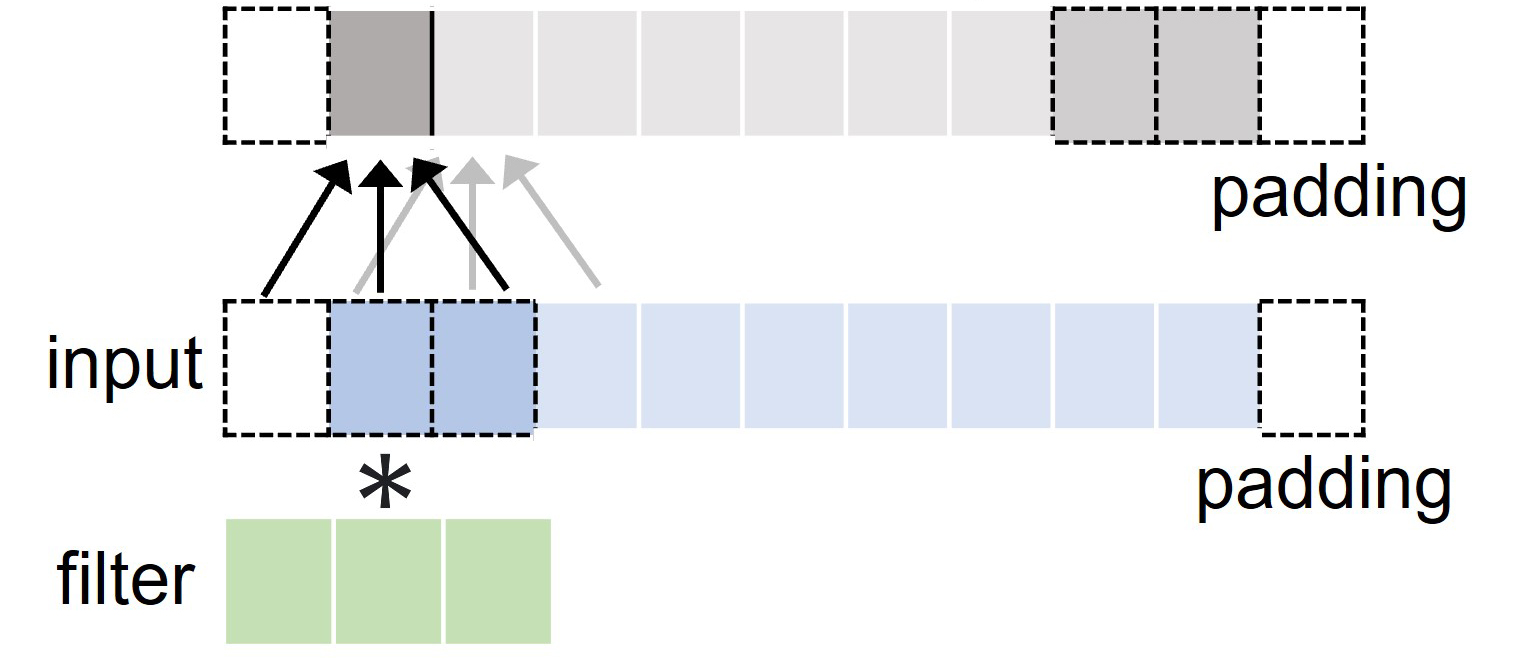} \\
      \footnotesize{(a)} & \footnotesize{(b)} \\
 \hspace{-1cm} 
      \includegraphics[height=0.22\textwidth]{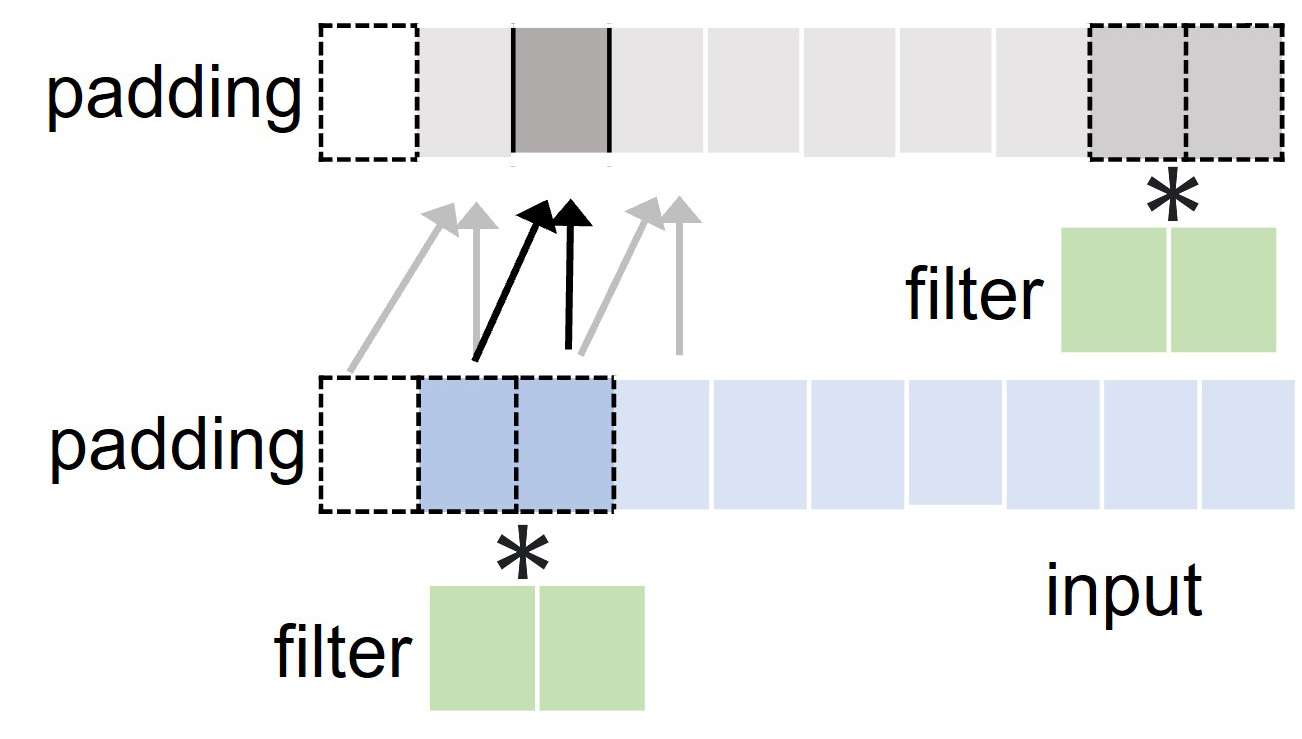} &
      \includegraphics[height=0.22\textwidth]{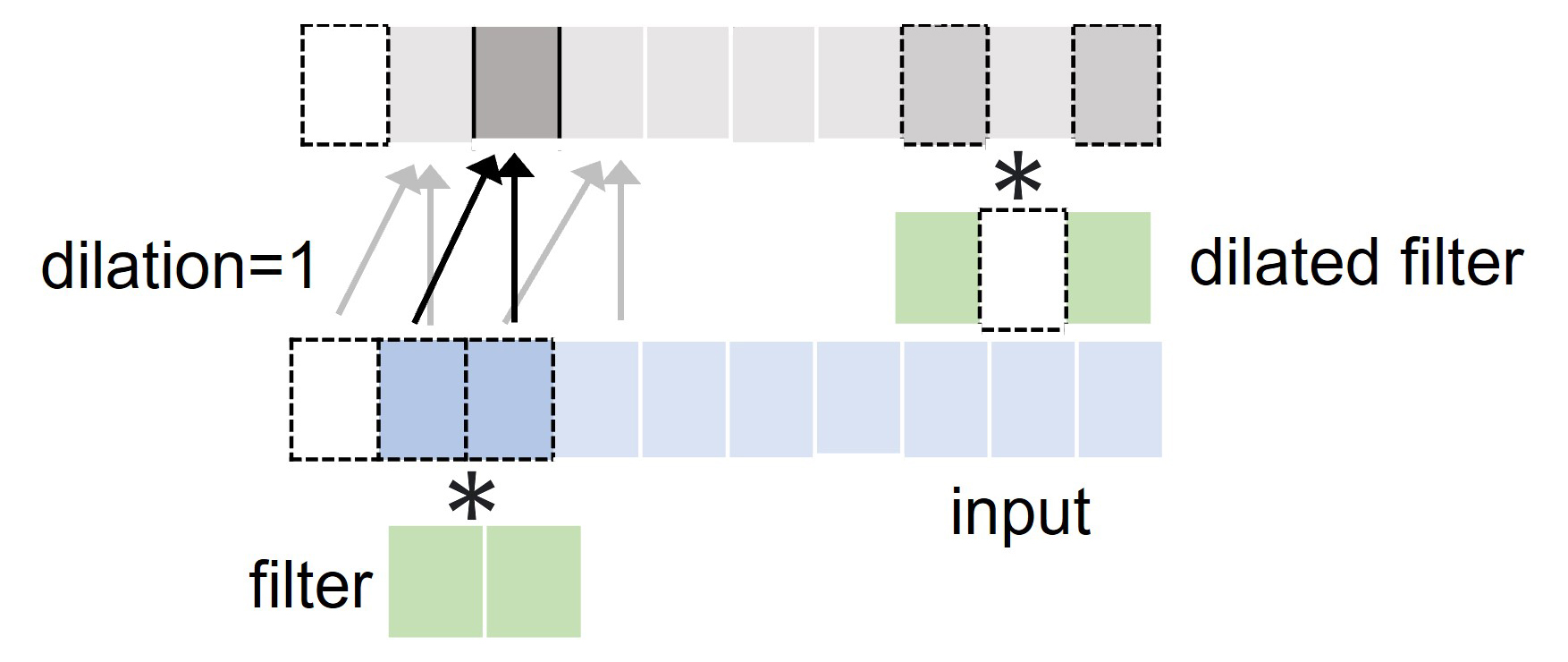} \\
      \footnotesize{(c)} & \footnotesize{(d)} \\
\hspace{5pt}
\end{tabular}
\captionof{figure}{Illustration of different types of CNN convolutions for a $9$-length time-series: (a) standard (vanilla) convolution process by a $3$-length filter, (b) padded convolution, (c) casual convolution and (d) dilated-causal convolution by a $2$-length filter.}
\label{fig:convolution}
\end{figure}

In time-series forecast, a pivotal element of the CNN model is the convolution process—a filter sliding over input time-sequence for feature extraction as depicted in Fig. ~\ref{fig:convolution}. Equation ~\ref{eq:conv_vanilla} defines the standard convolution operation ($C_{\text{vanilla}}$) as a dot product between a filter ($F$) and a time-sequence ($S$) of lengths $k$ and $l$ respectively. 
Because of the weight-sharing property of a CNN model, the entire time sequence is repeatedly convolved by the same filter.
Fig. ~\ref{fig:convolution} (a) illustrates this vanilla convolution using a $3$-step filter on a $9$-step sequence with a stride of $1$. Convolution results in a reduced-length sequence of $7$ steps ($l - k + 1 = 9 - 3 + 1 = 7$), which leads to information loss. To prevent this loss, \textit{padded} convolution ($C_{\text{padded}}$) operates on an extended sequence $S'$, created by padding zeros to the edges of $S$ [see Equation ~\ref{eq:conv_padded} and Fig. ~\ref{fig:convolution} (b)]. This ensures the input and convolved time sequences have the same length.


\begin{align}
C_{vanilla} &:=  F.S_{t:t+k}\; |\; \forall t \in [1, l-k] \label{eq:conv_vanilla} \\
C_{padded}  &:=  F.S'_{t:t+k}\; |\; \forall t \in [0, l-k+1] \label{eq:conv_padded}\\
C_{casual}  &:=  F.S'_{t-k+1:t}\; |\; \forall t \in [1, l] \label{eq:conv_casual}\\
C_{dilated-casual}  &:=  F'.S'_{t-(k+d-1)+1:t}\; |\; \forall t \in [1, l] \label{eq:conv_dilatedcasual}
\end{align}

\vspace{1em}

Casual convolution ($C_{casual}$) is often a preferable algorithm in time-series forecasting to avoid information leakage from the future ~\cite{CNN_casualdilated}.
It precedes the input sequence with $k$-1 zeroes before convolution, as illustrated in Fig.~\ref{fig:convolution} (c). Casual convolution would require a large filter or a deep network when it needs to examine a large number of past observations. Dilated Convolution overcomes this drawback by introducing a dilation factor $d$ into the filter so that convolution can be applied on an input area larger than the original filter size, i.e., capturing larger contextual information. The modified filter $F^{'}$ is obtained by dilating the original filter with ($d-1$) zeros as shown in Fig. ~\ref{fig:convolution} (d). 
Dilation is performed using $d=2$. The positions of zero correspond to the skipped timesteps in the input sequence. 
Both the casual (in Equation ~\ref{eq:conv_casual}) and dilated casual convolutions ($C_{dilated-casual}$) in Equation ~\ref{eq:conv_dilatedcasual} output a sequence with the same length of the input. In the following, we introduce three CNN model configurations to perform the multivariate scour forecast:



1) Vanilla Convolutional Network (VCN)\label{sec:vanillacnn}:
The convolution block of the VCN model comprises one layer of standard convolution (from Equation \ref{eq:conv_vanilla}) and one layer of nonlinear activation function, i.e., Rectified Liner Unit (ReLU). Two identical convolution blocks are stacked together as a feature extractor that precedes a flattened layer, a drop-out layer, and a fully connected dense layer. 
Finally, a reshape layer is added to ensure that the predicted time sequence contains the same number of sensor variables as input.

\textit{2) Fully Convolutional Network (FCN)}:
A FCN ensures that the length of the convolved sequence is the same as the input. Here, the zero-padded convolution from Equation ~\ref{eq:conv_padded} with a stride value of $1$ is used. We followed the structure adopted by Shelhamer et al. ~\cite{FCN1} for the FCN in this study, where the main block consists of one convolution layer, one batch normalization (BN) layer, and one ReLU layer. BN layer ~\cite{BN} ensures fast convergence and improves generalisation performance. The rest of the setup remains the same as the VCN. 

\textit{3) Dilated Causal Network (DCN)}:
As shown in  Equation ~\ref{eq:conv_dilatedcasual} and Fig. \ref{fig:convolution} (d), dilation factors of $1$ and $2$ are applied in two convolution layers. This enables the CNN model to perform causal convolution at a fine-grained level on the input sequence and a coarse-grained level on the convolved sequence (from the lower layer). Each convolution layer is followed by one BN layer and one ReLU layer.
Model configuration nomenclatures for a temporal CNN are encoded as follows (with $k$ and $F$ being the filter size and no. of filters of the layers, respectively):
\begin{equation}\label{eq:encoding_config_CNN}
    [model]-[k_1]-[F_1]-[k_2]-[F_2]-[dropout]
\nonumber
\end{equation}

where the model is different CNN variants introduced, $k_1$ and $k_2$ are the kernel size, $F_1$ and $F_2$ are the hidden size of the dense layer, and dropout is the dropout rate applied to the model.
\subsection{Model Evaluation Criteria}\label{sec:loss}
We used Mean Square Error (MSE) on the \textit{Sonar} readings to train the scour forecast model. For $S$ no. of samples in a batch (32) with $x_{i,t}$ and $\hat{x}_{i,t}$ representing the actual and predicted values of \textit{Sonar} feature $X_{\textit{sonar}}$ at timestep $t$ for the $i^{th}$ sample, the training loss is computed as follows. 
\begin{align} 
MSE = \frac{1}{S*w_{out}}\sum_{i=1}^{S}\sum_{t=1}^{w_{out}}(x_{i,t}-\hat{x}_{i,t})^2, x_{i,t} \in X_{\textit{sonar}}
\label{Eq:MSE}
\end{align}
We evaluated the scour prediction performance of a model using the metric MAE on the \textit{Sonar} readings. For $S$ number of samples in a batch with $x_{i,t}$ and $\hat{x}_{i,t}$ representing the actual and predicted values of \textit{Sonar} feature $X_{\textit{sonar}}$ at timestep $t$ for the $i^{th}$ sample, the performance metric, MAE is computed as follows. 
\begin{align} 
MAE = \frac{1}{S*w_{out}}\sum_{i=1}^{S}\sum_{t=1}^{w_{out}}|x_{i,t}-\hat{x}_{i,t}|, x_{i,t} \in X_{\textit{sonar}}
\end{align}

We used the hold-out method \cite{cv1,cv2} for evaluating the performance of the models. In the same unit as sensor readings, MAEs are reported in foot elevation throughout.
\subsection{Hyperparameter Tuning} \label{sec:random_policy}
 DL model optimisation and HP tuning using grid-search is computationally expensive \cite{LSTM1}. Following the grid-search approach in ~\cite{phase2}, the performance of all possible HP combinations was analysed by running each configuration for a large number of iterations. 
To explore a more computationally efficient alternative to this approach, we investigated a random-search strategy 
based on the following three possible policies to select the best model configurations. In this approach, a sample of HP combinations with $50$\%, $35$\%, and $67$\% sample size($s$) out of all possible combinations are randomly selected in a number of trials ($t=20$). 

The grid-search results were used as a benchmark for the "true" performance of each configuration and to evaluate the outcome of these random search strategies.

\begin{enumerate}

\item \textit{meanMAE Policy}: Model configurations are ranked based on the average MAE values across random trials. Lower average MAE values lead to higher rankings; this is irrespective of the number of times a model configuration appears across all trials (=$f$).

\item \textit {medianMAE Policy}: Model configurations are ranked by taking the median of the MAE values across the trials. Models with smaller median MAE would receive a higher rank than the model with larger median values; similar to mean MAE policy, this is irrespective of the number of times a configuration appears across all trials. 

\item \textit{Bagging (Bootstrap Aggregation) Policy} \cite{Bagging1}:
In this policy, top $k$ configurations with the lowest MAE values are selected on each trial. Considering $f_{\textit{topk}}$ as the number of times a particular configuration is ranked as top-$k$ across $t$ trials, $k$ models (out of possible $k*t$ models) are finally selected based on $f_{\textit{topk}}$; model configurations with more frequent appearances in the top-$k$ (i.e., larger $f_{\textit{topk}}$) receive higher ranks. 

\end{enumerate}

Finally, to observe how well the trained models generalize on unseen data, we generated prediction plots on the test datasets for the top-5 models. 
These top-5 models were selected after careful observations of the mean and standard deviation of MAE scores from both the random and grid-search policies. The prediction plots evaluate the performance of DL models in recognizing the scour (and filling) trends across time.

\subsection{Experiments}
Table \ref{table:HP-tuning} provides a summary of experiments we conducted to address the research questions outlined in Section ~\ref{sec:intro}. For all experiments except for Feature Combination Analysis, \texttt{sN} and \texttt{sT} are used as input features.

\begin{table}[h]
\footnotesize
\vspace{10pt}
\caption{Description of the experiments and their corresponding datasets, hyperparameters, and tunning values.}
\begin{tabular}{|l|l|l|l|l|}
\hline
Experiment & Dataset & HPs & Tuning Values \\
\hline
 1. HP & Alaska 742 & ($w_{in}$, $w_{out}$) & \{(168, 168), (336, 168), (720,168)\} \\  
  Tunning Strategy & & units & \{32, 64, 128\} \\ 
  & & models & \{ss\}\\ 
  & & dropouts & \{0.0, 0.2\} \\ \hline
 2. LSTM  & Alaska 230,539, 742 & ($w_{in}$, $w_{out}$) &  \{(168, 168), (336, 168), (720,168)\} \\  
 Evaluation &  & units & \{32, 64, 128, 256\} \\  
 & Oregon Trask, Luckiamute  & models& \{ss, ss2, fb\} \\
 & & dropouts & \{0, 0.2\} \\
 \hline
 & \multirow{4}{*}{\begin{tabular}[c]{@{}l@{}}Alaska 230, 539, 742\\  
 Oregon Trask, Luckiamute\end{tabular}} & ($w_{in}$, $w_{out}$) & \{(168, 168), (336, 168), (720,168)\} \\  
 &  & layer-1 filter length: $k_1$ & \{3, 5, 7\}\\ 
3. CNN   &  &  layer-2 filter length: $k_2$ & \{5, 7, 11\}\\ 
Evaluations &  &  layer-1 filters: $F_1$ & \{64, 128, 256\}\\ 
 &   & layer-2 filters: $F_2$ & \{128, 512, 1024\}\\ 
 & &  models & \{vcn, dcn, fcn\} \\ 
 &  & dropout & \{0.0, 0.2\} \\ \hline
4. Feature  & Alaska 230, 539 & features & \{sN, sT, dV, dC, sNsT, sNdC, sNdV, \\
Combinations & Oregon Luckiamute & & sTdV, sTdC, sNsTdV, sNsTdC\}\\
\hline 
\end{tabular}
\label{table:HP-tuning}
\end{table}

\section{Results}\label{sec:exp_design}

\subsection{Evaluation of Random Policies for HP Tuning (Experiment 1)}\label{sec:random_policy_results}
We evaluated the HP selection policies on Alaska $742$ bridge, using the single-shot LSTM scour forecast models. 
Given the past $1$, $2$, $3$ weeks ($168$, $336$, $720$ hrs) observations (\textit{$w_{in}$}), we forecast hourly flow and bed elevation variation, aiming to predict scour for the upcoming week (\textit{$w_{out}$}). Here, the full search space comprises $18$ distinct HP configurations as per Table \ref{table:HP-tuning}. 
TALOS's \textit{uniform mersenne} strategy ~\cite{RS3} is used for random selection of HPs. 

\begin{figure}[htbp]
    \centering
    \begin{subfigure}[t]{0.45\textwidth}
        \centering
        \includegraphics[width=1\textwidth]{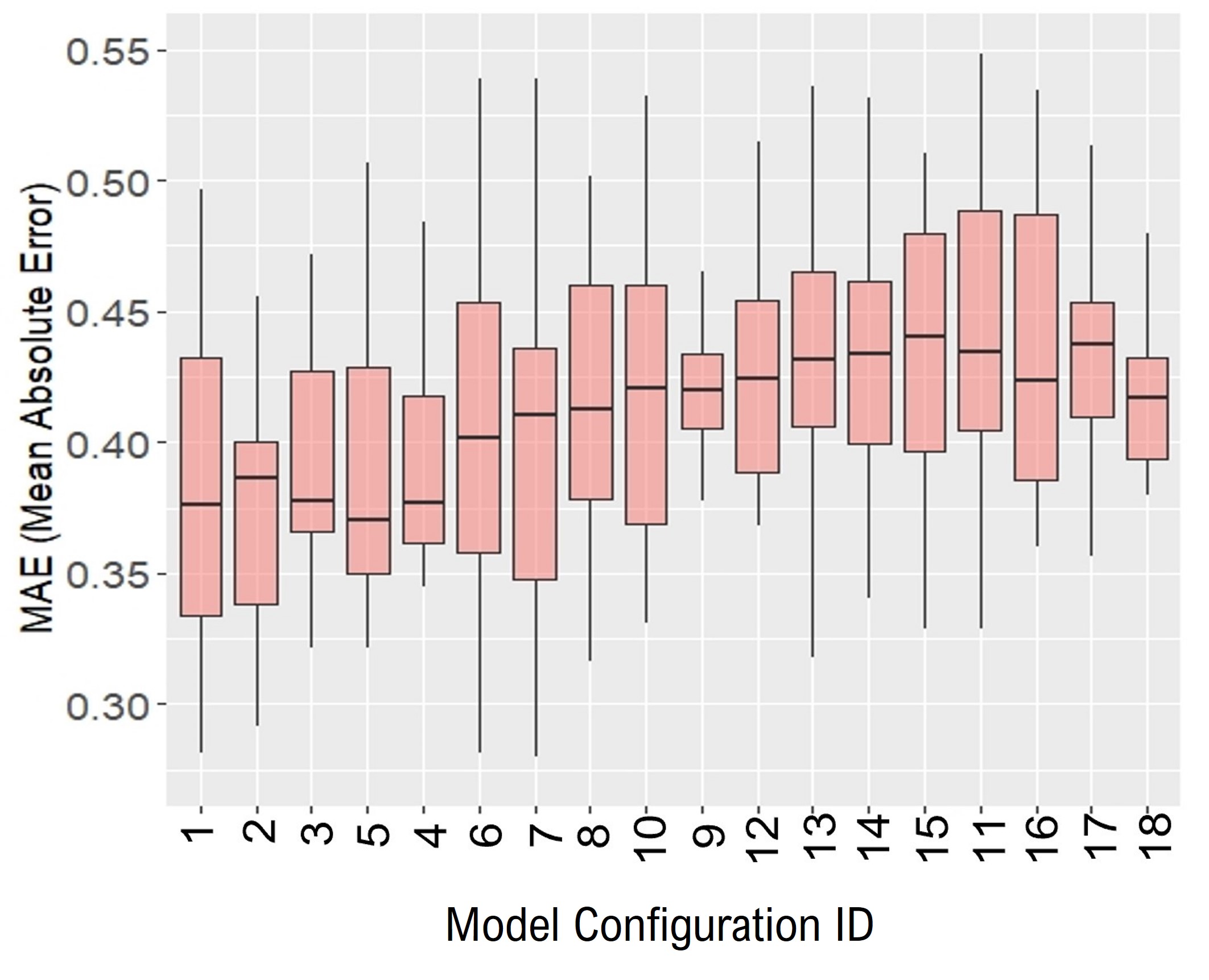}
        \subcaption{}
    \end{subfigure}
    \hfill
    \begin{subfigure}[t]{0.45\textwidth}
        \centering
        \includegraphics[width=1.0\textwidth]{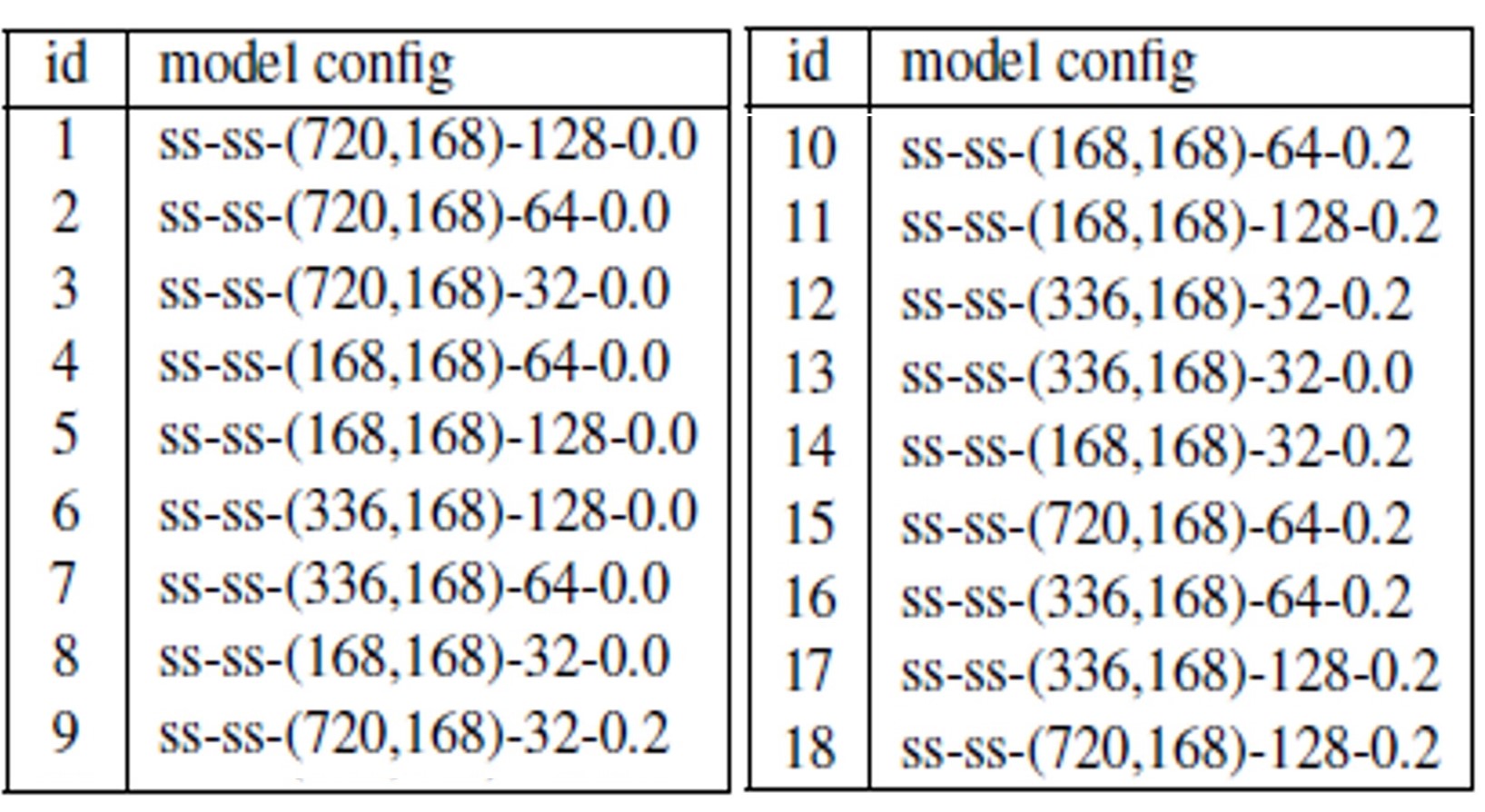}
        \subcaption{}
    \end{subfigure}

    \vspace{20pt}
    \begin{subfigure}[t]{\textwidth}
        \centering
        \includegraphics[width=0.8\textwidth]{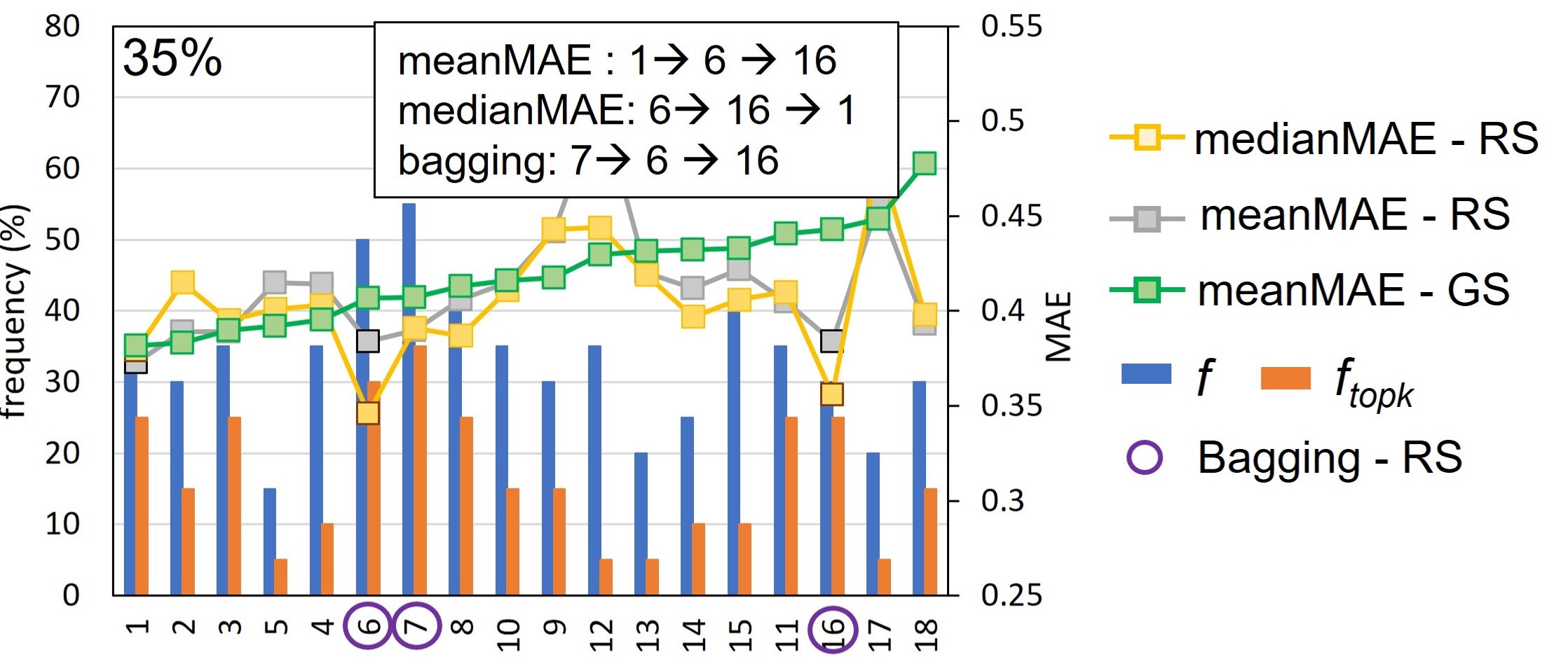}
        \subcaption{}
    \end{subfigure}
    \vspace{20pt}

    \begin{subfigure}[t]{0.45\textwidth}
        \centering
        \includegraphics[width=1.2\textwidth]{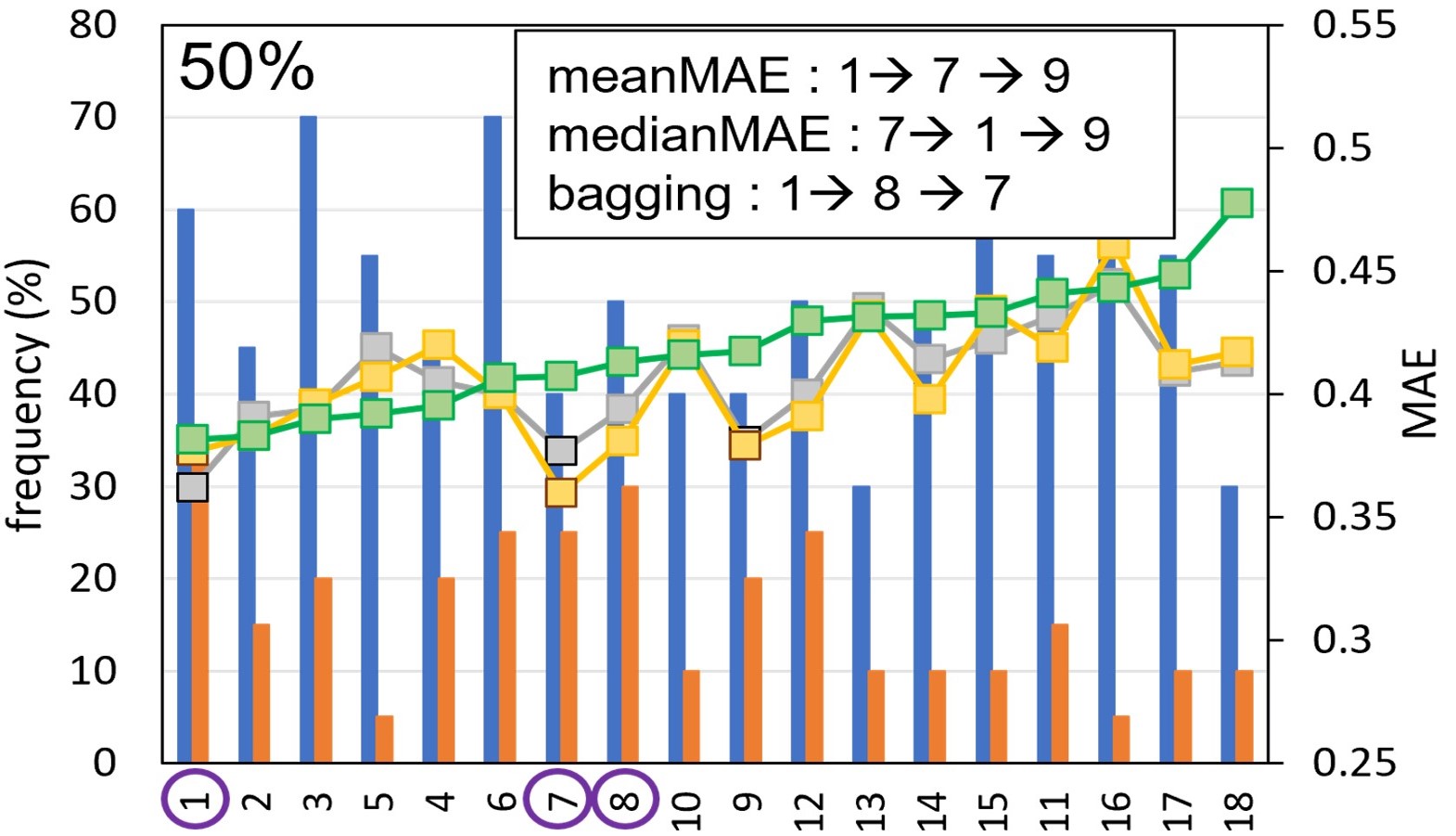}
        \subcaption{}
    \end{subfigure}
    \hfill
    \begin{subfigure}[t]{0.45\textwidth}
        \centering
        \includegraphics[width=1.2\textwidth]{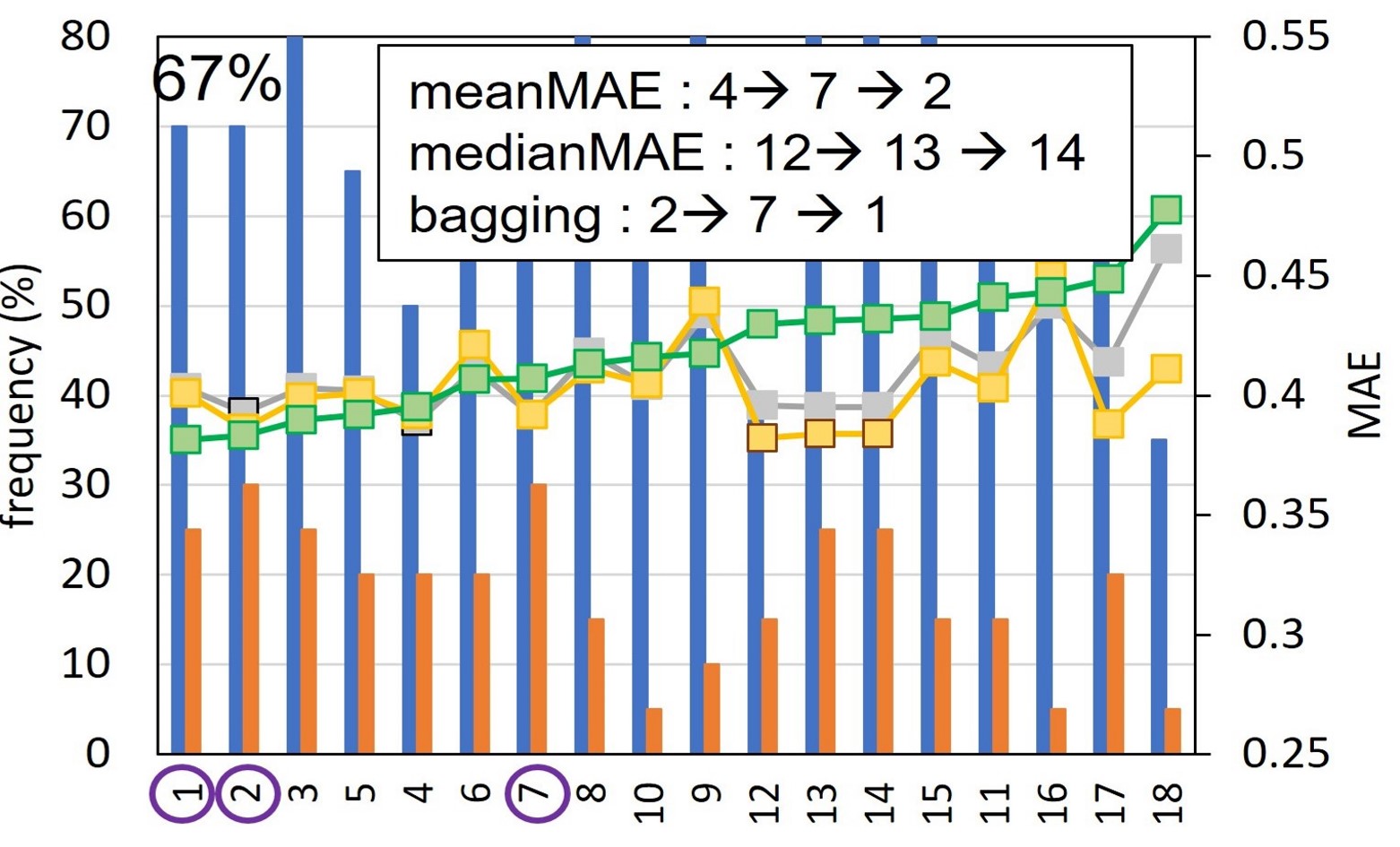}
        \subcaption{}
    \end{subfigure}
    \vspace{20pt}
    \caption{Comparison of random-search (RS) policies with grid-search (GS) for HP tuning: Ranked LSTM model configurations based on grid-search (a), showing all possible HP combinations for $18$ models (b), frequency, top-$k$ frequency, mean MAE and median MAE of HP configurations appearing in random-search iterations for (c) $35$\%, (d) $50$\%, (e) $67$\% sample size (top-3 models picked based on each policy are shown at top.}
    \label{Fig:Bardiagrams_GSRS}
\end{figure}

Fig. ~\ref{Fig:Bardiagrams_GSRS} demonstrates the results of HP tunning using grid-search and random-search policies for the $18$ scour forecast models. Top-3 models with the smallest average MAE from grid-search (GS) served as benchmarks when evaluating random-search (RS) policies. Since the distribution of MAE values is not heavily skewed as observed in the boxplots, the mean values are used to rank the models in Fig. ~\ref{Fig:Bardiagrams_GSRS} (a). It is noted that higher ranked models (top-$5$) would remain the same with median criteria. Note that all the reported MAE values throughout the paper are in m (of elevation).

The meanMAE and medianMAE random-search policies show similar trends in lower sample sizes (35\% and 50\%), but as expected, they deviate by increasing the sample size.  With a growing sample size, the mean-MAE estimations seem to converge towards grid-search estimates. 
The bagging policy also fails to find the top-ranked models when using small sampling sizes; however, with increasing sample size (67\%), this method shows to be comparatively more robust and less sensitive to the poor-performing models compared to the meanMAE and medianMAE policies (e.g. models 12 to 18 show low ranking based on bagging but rank high in mean and median policies). Furthermore, the meanMAE and bagging policies seem to agree with 2 out of 3 top selected configurations, whereas the medianMAE policy fails to identify any of the top-performing models [Fig. ~\ref{Fig:Bardiagrams_GSRS} (e)].

\subsection{LSTM Evaluation (Experiment 2)}\label{sec:LSTM_results}
The three variants of scour forecast LSTM models were evaluated to explore performance on 1) the recent years of collected sensor data from Alaska case study bridges, $230$, $539$, and $742$ and 2) Oregon case study bridges, Trask and Luckiamute, with drastically different scour behaviour. 
Alaska and Oregon LSTM models were trained using the hold-out method, by partitioning the data into training, validation, and test subsets, with a splitting ratio (\%) of $70$-$20$-$10$ and $60$-$20$-$20$, respectively. 

\begin{table}[!ht]
\centering
\vspace{10pt}
\scriptsize
\caption{Top-$5$ LSTM models selected by random-search meanMAE policy for Alaska bridge $230$, $539$ and $742$ and two Oregon bridges, Trask and Luckiamute.}
\begin{tabular}{@{}|c|c|c|l|c|c|c|l|@{}}
\hline
Bridge & Rank & Models & meanMAE \\
\hline
\multirow{5}{*}{\rotatebox[origin=c]{90}{230}} 
  & 1 & ss-(336,168)-128-0 & 0.078 \\  
  & 2 & ss2-(336,168)-32-0 & 0.083 \\
  & 3  & ss-(336,168)-32-0.2 & 0.083 \\
  & 4 & ss-(720,168)-64-0.2 & 0.083 \\
  & 5 & ss-(336,168)-32-0 & 0.083 \\
\hline
\multirow{5}{*}{\rotatebox[origin=c]{90}{539}} 
  & 1 & ss2-(720,168)-128-0 & 0.243 \\
  & 2 & ss-(720,168)-64-0 & 0.253 \\
  & 3 & ss-(168,168)-128-0 & 0.259 \\
  & 4 & ss2-(168,168)-64-0 & 0.266 \\
 & 5 & ss2-(720,168)-32-0 & 0.27 \\
\hline
\multirow{5}{*}{\rotatebox[origin=c]{90}{742}} 
  & 1 & ss-(168,168)-128-0 & 0.41 \\
  & 2 & ss-(336,168)-64-0.2 & 0.413 \\
  & 3 & ss-(336,168)-128-0 & 0.419 \\
  & 4 & ss-(336,168)-32-0 & 0.424 \\
  & 5 & ss-(720,168)-32-0.2 & 0.431 \\
\hline
\multirow{5}{*}{\rotatebox[origin=c]{90}{Trask}} 
  & 1 & ss-(168,168)-32-0 & 1.573 \\
  & 2 & ss-(168,168)-64-0.2 & 1.615 \\
  & 3 & ss-(168,168)-32-0.2 & 1.62 \\
  & 4 & ss-(168,168)-64-0 & 1.638 \\
  & 5 & ss-(336,168)-32-0.2 & 1.648 \\
\hline
\multirow{5}{*}{\rotatebox[origin=c]{90}{Luckiamute}} 
  & 1 & ss-(720,168)-32-0 & 0.249 \\
  & 2 & ss-(720,168)-64-0 & 0.251 \\
  & 3 & ss-(720,168)-32-0.2 & 0.253 \\
  & 4 & ss-(720,168)-128-0 & 0.257 \\
  & 5 & ss2-(720,168)-32-0.2 & 0.261\\
 \hline
\end{tabular}
\label{table:LSTM_Top5}
\end{table}

The HP configurations (see Table \ref{table:HP-tuning}) are optimized based on random-search method with meanMAE policy using \%67 sample size. The top-5 LSTM model configurations for each bridge are presented in Table \ref{table:LSTM_Top5}. The best-performing models achieve mean MAE values ranging from 0.078 to 0.410 ft (0.12 m) for Alaska bridges and 0.249 to 1.573 ft (0.48m) for Oregon bridges. 

The Single-Shot (ss) variant shows consistently better performance than Feedback (fb) variant for all bridges. Models with larger number of past observations, e.g., $w_{in}=720$, obtain smaller mean errors and appear as higher rank models. An input width 2 or 3 times the output/target width (forecast window) seems to yield the most accurate scour predictions.

\subsubsection{Scour Forecast for Alaska Bridges}\label{sec:LSTM_results_Alaska}
Fig. ~\ref{Fig:230_prediction} and Fig. ~\ref{Fig:539_prediction} present the predicted values of \textit{Sonar} (bed elevation) and \textit{Stage} (water elevation) by the top-5 models for bridge $230$ and $539$ across the test dataset, between July-$2021$ to Sep-$2021$. The performance of models is assessed based on bed elevation (scour depth) predictions. 

The models show reasonable prediction accuracy and are able to capture the scour and filling trends a week in advance throughout the test (unseen) period of the data.

\begin{table*}[htbp]
\centering
\begin{tabular}{c}
      \resizebox{\columnwidth}{!}{\includegraphics{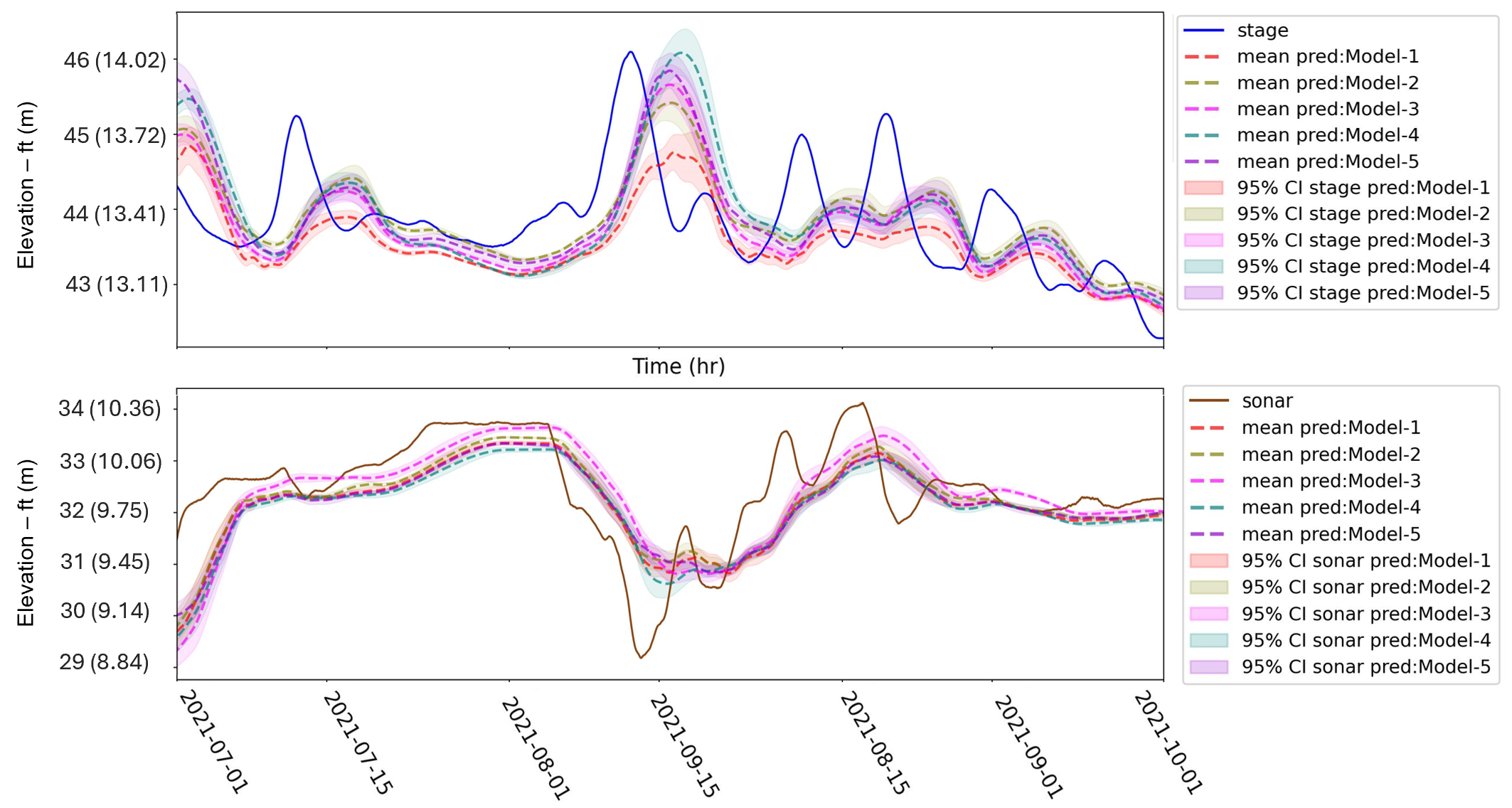}}
\end{tabular}
\captionof{figure}{Stage (top) and Sonar (bottom) predictions of top-$5$ LSTM models selected by the random-search (meanMAE policy), mean and 95\% confidence intervals over the test dataset -  Alaska bridge $230$.}
\label{Fig:230_prediction}
\vspace{10pt}
\end{table*}

\begin{table}[htbp]
\centering
\begin{tabular}{c}
      \resizebox{\columnwidth}{!}{\includegraphics{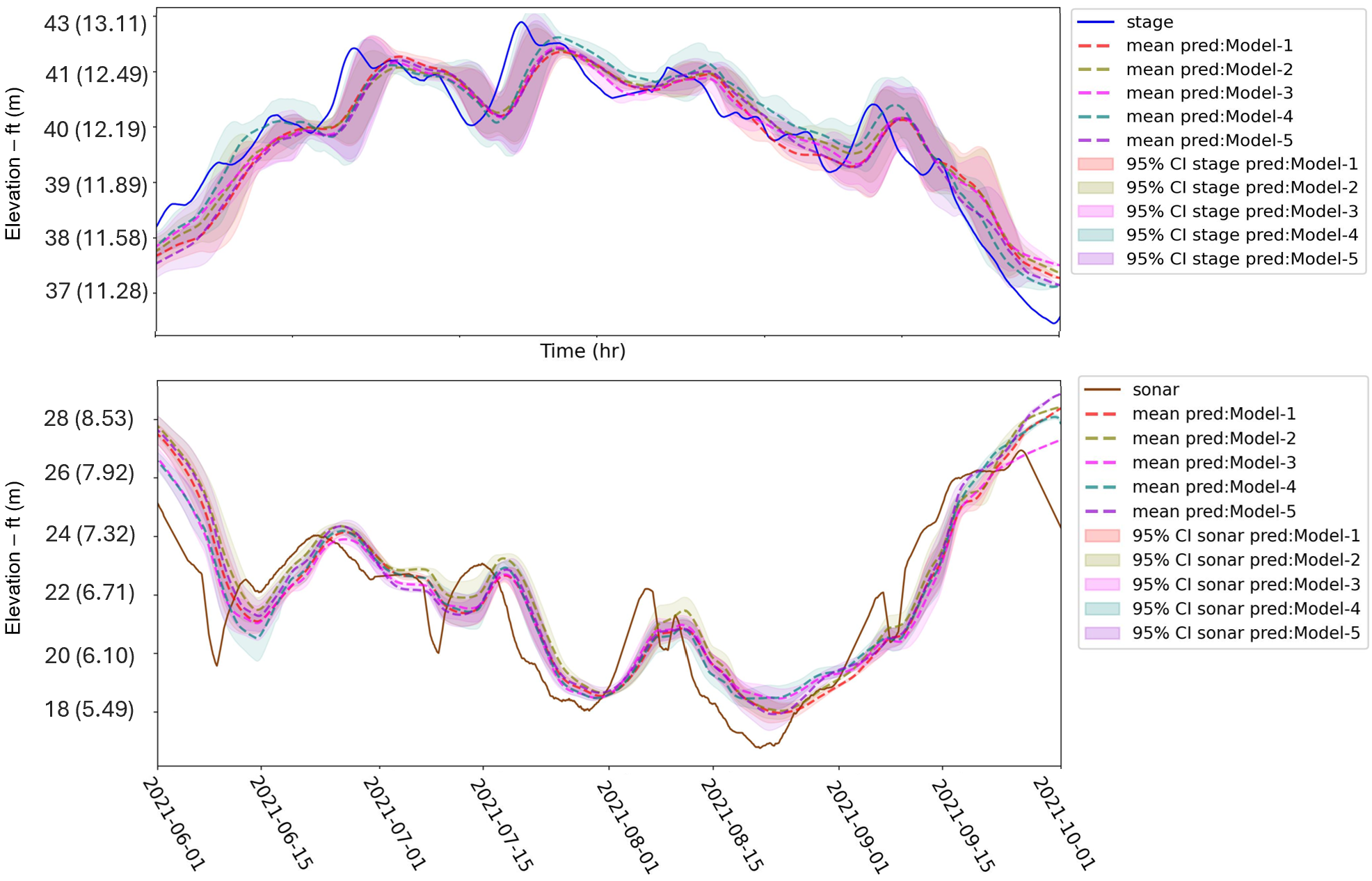}}
\end{tabular}
\captionof{figure}{Stage (top) and Sonar (bottom) predictions of top-$5$ LSTM models selected by the random-search (bagging policy), mean and 95\% confidence intervals over the test dataset -  Alaska bridge $539$.}
\label{Fig:539_prediction}
\end{table}

For bridge $230$, during the peak time of scour (August), the maximum error between the predicted and the actual value of scour is around 2ft (0.6m), whereas for the $539$ bridge, it is around 1ft (0.3m). The inverse relation of the scour trend with the stage-water is well captured by the LSTM models with negotiable lags for all bridges. 

\begin{table}[htbp]
\centering
\begin{tabular}{c}
      \resizebox{\columnwidth}{!}{\includegraphics{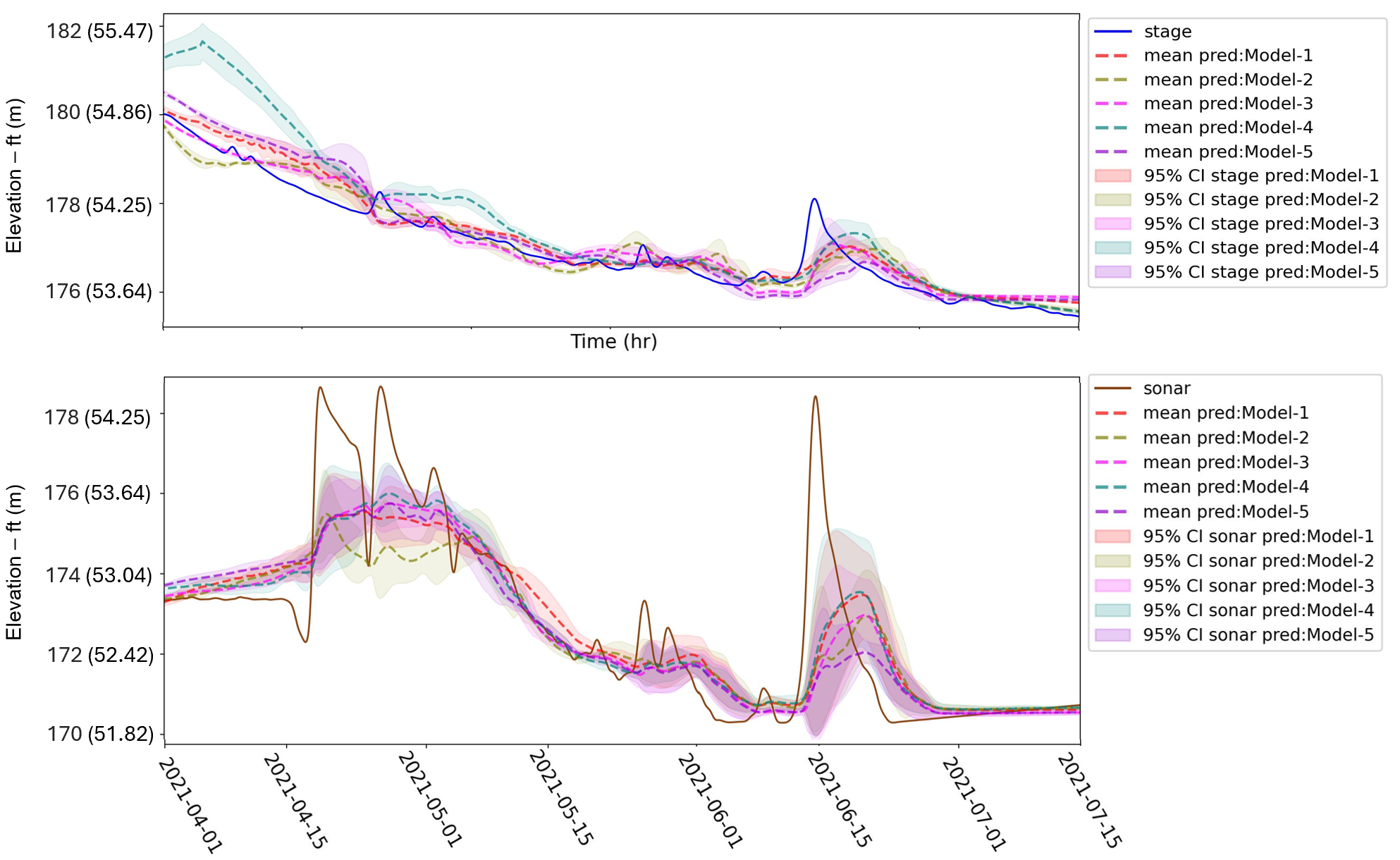}}
\end{tabular}
\captionof{figure}{Stage (top) and Sonar (bottom) predictions of top-$5$ models selected by the meanMAE policy, showing the mean and 95\% confidence interval over the test dataset - Luckiamute bridge.}
\vspace{10pt}
\label{Fig:Luckiamute_prediction}
\end{table}

\begin{table}[htbp]
\centering
\begin{tabular}{c}
      \resizebox{\columnwidth}{!}{\includegraphics{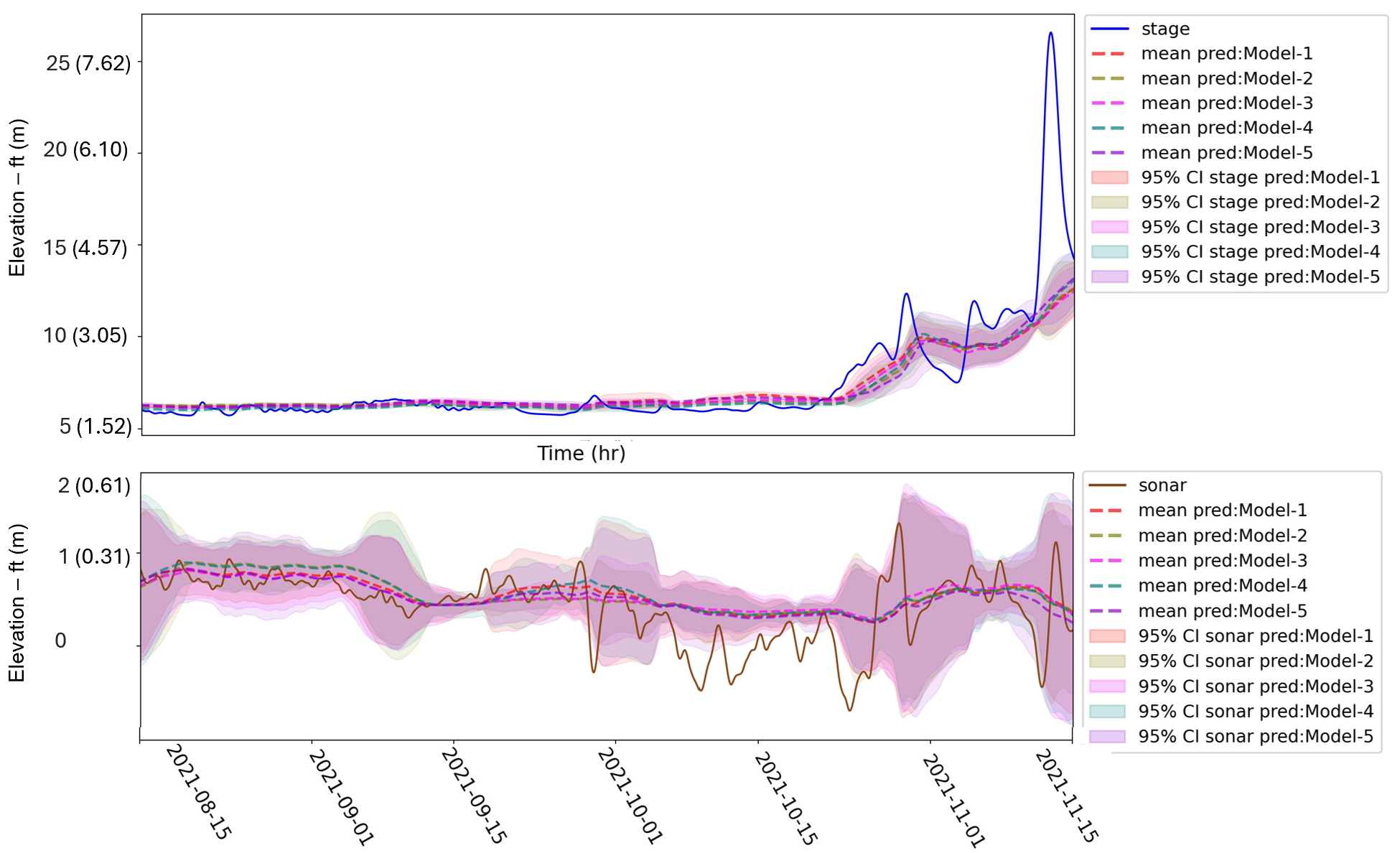}}
\end{tabular}
\captionof{figure}{\footnotesize{Stage (top) and Sonar (bottom) predictions of top-$5$ models selected by the meanMAE policy, showing mean and 95\% confidence interval over the test dataset - Trask bridge.}}
\label{Fig:Trask_prediction}
\end{table}

\subsubsection{Scour Forecast for Oregon Bridges}\label{sec:LSTM_results_Oregon}
 Stage and Sonar prediction of top-5 models over the test dataset is presented in Fig. ~\ref{Fig:Luckiamute_prediction} and Fig. ~\ref{Fig:Trask_prediction} for Luckiamute and Trask bridges, respectively. The test data is selected for the period between April-$2021$ to July-$2021$ and August-$2021$ to Nov-$2021$, for Luckiamute and Trask, where significant scour and fillings are observed.


 For Oregon Luckiamute, some of the large spikes in bed elevation seem to be due to errors in reading, which were not picked up by pre-processing techniques (verified with Oregon USGS). Yet, LSTM models are able to identify the realistic trend of bed elevation variation.  For Trask which is a tidal river showing frequent daily fluctuations due to low and high tides no significant scour/filling is observed with less than 2ft (0.6m) change in bed level. The mean predictions fail to capture these frequent fluctuations, resulting in an underestimation of scour depth. However, the lower-bound values seem to capture maximum scour depths with some lag. Nevertheless, the Trask bed elevation variation is less than a meter and does not show significant scouring. 
 
 Overall, the LSTM models perform better in capturing peak values of scour and fill episodes for Alaska bridges compared to Oregon in our case studies. This can be attributed to the distinct scour process for the two Oregon bridges. Both Oregon bridges are over coastal rivers with soft slit sediments and bedforms (dunes). Sonar and Stage do not show the typical reverse correlation as in Alaskan bridge with riverine flows and consistent seasonal patterns ~\cite{Sheppard1993,arneson2013evaluating}. In addition, the data availability in these locations is significantly less than in Alaska (3 years of historical data for Oregon versus 15 years for Alaska). 

\subsubsection{Sequential Training on Oregon Data}\label{sec:sequentialtraining}
\begin{figure}[htbp]
\centering
\includegraphics[width=.5\textwidth]{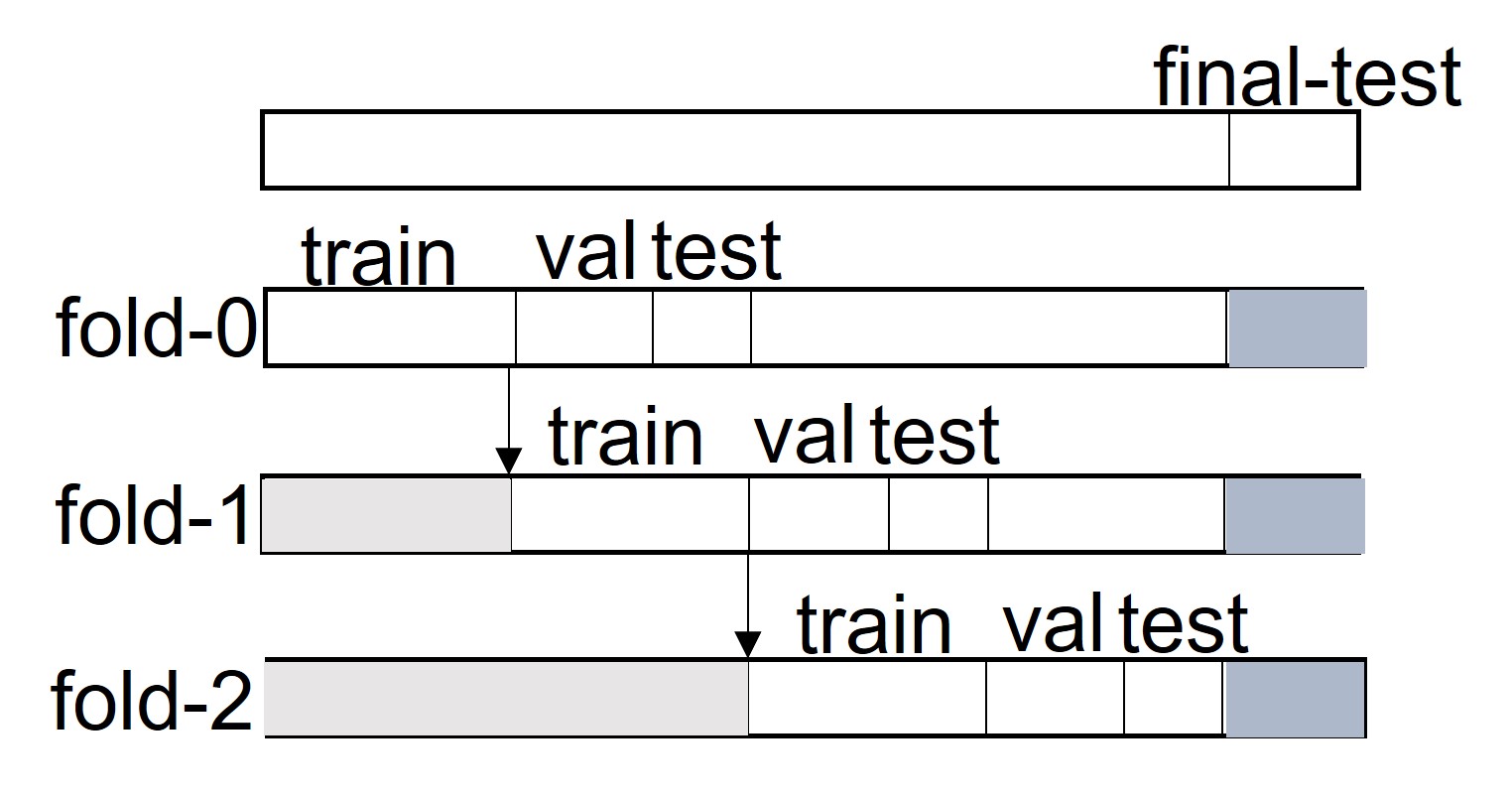}
\caption{Illustration of 3-fold sequential training and validation.}
\label{Fig:SQTraining}
\vspace{10pt}
\end{figure}

The hold-out training method is sufficient for the LSTM models to learn the underlying patterns of scour/flow characteristics at coastal rivers such as Trask and Luckimaute, leading to poor generalization.  Cross-validation \cite{cv1,cv2} could be an alternative, however it is computationally expensive. Therefore, motivated by transfer learning \cite{transferleraning}, we implemented a ``Sequential Training" technique as shown in Fig. \ref{Fig:SQTraining} to improve the forecast accuracy for Oregon bridges. For this experiment, we used the top-3 LSTM model configurations from Section  \ref{sec:LSTM_results_Oregon}.

The major part ($90$\%) of the data is repeatedly split into 3 Folds and 5 Folds, with a small portion ($10$\%) set aside for the final evaluation. Each of the folds is further divided into three parts: train, validation, and test. The model is trained using the folds in a sequential manner, i.e., the model trained (with its learned weights and biases) in a fold is going to be fine-tuned using the newer train data from the next fold. There is no overlap across the training data between two adjacent folds. This data division and the transfer of ``learning" between folds saves significantly on training time and expected to improve the performance of DL models.

\begin{figure}[htb]
\vspace{0pt}
\centering
\includegraphics[width=\textwidth]{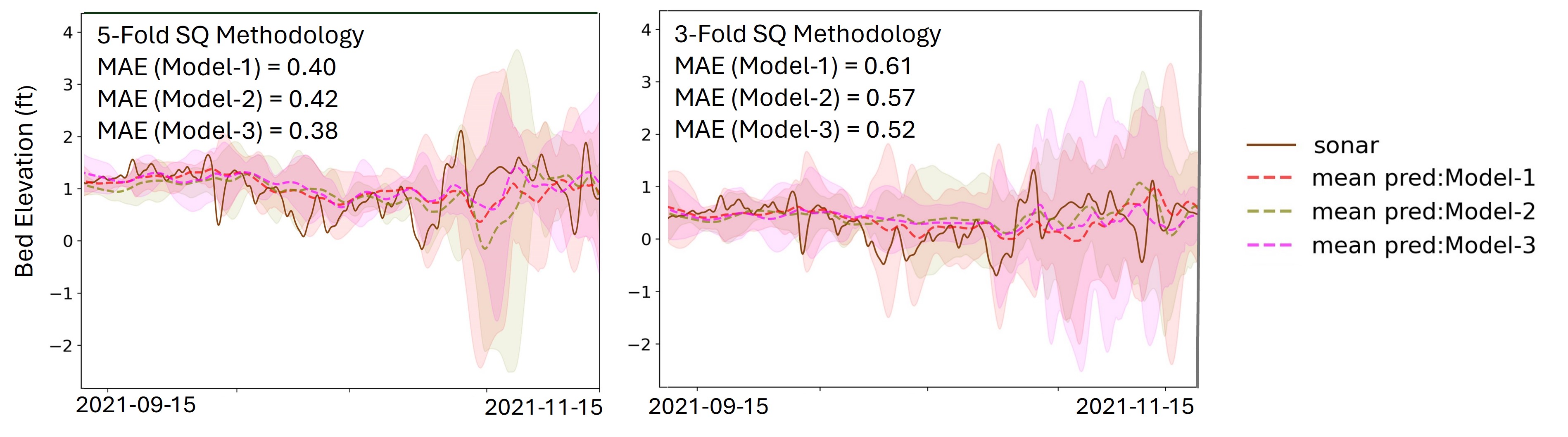}
\caption{Bed elevation (Sonar) predictions using top LSTM models with 5-Fold and 3-Fold sequential training method for Oregon Trask bridge.}
\label{Fig:SQ_TraskPrediction}
\vspace{12pt}
\end{figure}

Table \ref{table:SQ_OregonResults} presents the MAE values on Sonar, and Fig. \ref{Fig:SQ_TraskPrediction} shows the time-series forecast over the final-test data using the 3-Fold and 5-Fold sequentially-trained LSTM models for Trask bridge. It is observed that the prediction error is reduced using the Sequential Training method.

\begin{table}[htb]
\vspace{15pt}
\caption{MAE of 3-Fold and 5-Fold sequentially-trained LSTM models - Trask bridge.}
\resizebox{135mm}{!}{
\begin{tabular}{|c|c|lll|lll|lll|}
\hline
 &  & \multicolumn{3}{c|}{model-1} & \multicolumn{3}{c|}{model-2} & \multicolumn{3}{c|}{model-3} \\
 &  & \multicolumn{3}{c|}{ss-(168,168)-32-0} & \multicolumn{3}{c|}{ss-(168,168)-64-0.2} & \multicolumn{3}{c|}{ss-(168,168)-32-0.2 } \\ \hline
SQ & fold & val & test & final\_test & val & test & final\_test & val & test & final\_test \\
\hline
\multirow{1}{*}{Hold-out}
 & N/A & 1.57 & N/A & 0.66 & 1.62 & N/A & 0.72 & 1.62 & N/A & 0.71 \\
\hline
\multirow{3}{*}{3-Fold} & 0 & 0.79 & 0.91 & 0.88 & 0.87 & 1.03 & 0.89 & 0.67 & 0.95 & 0.76 \\
 & 1 & 0.45 & 0.74 & 0.59 & 0.47 & 0.82 & 0.53 & 0.51 & 0.81 & 0.56 \\
 & 2 & 2.35 & 0.36 & 0.61 & 2.21 & 0.38 & 0.57 & 2.49 & 0.31 & 0.62 \\
 \hline
\multirow{5}{*}{5-Fold} & 0 & 0.42 & 0.63 & 0.66 & 0.42 & 0.69 & 0.79 & 0.37 & 0.67 & 0.84 \\
 & 1 & 0.95 & 0.62 & 0.66 & 1.01 & 0.69 & 0.76 & 1.17 & 0.60 & 0.71 \\
 & 2 & 0.38 & 0.35 & 0.53 & 0.60 & 0.34 & 0.60 & 0.64 & 0.35 & 0.62 \\
 & 3 & 1.15 & 2.81 & 0.64 & 1.21 & 3.17 & 0.59 & 1.25 & 2.90 & 0.64 \\
 & 4 & 0.36 & 0.14 & 0.40 & 0.60 & 0.21 & 0.42 & 0.50 & 0.14 & 0.38\\
 \hline
\end{tabular}
}
\label{table:SQ_OregonResults}
\end{table}

\subsection{CNN Evaluation (Experiment 3)}\label{sec:CNN_results}
Three different variations of CNN models, i.e., VCN, DCN, and FCN as described in Section \ref{sec:CNN_models} are trained for bridge $230$ and $539$ in Alaska and Luckiamute bridge in Oregon. \textit{Stage} and \textit{Sonar} are considered as both the input and target features in these models. 
Table \ref{table:HP-tuning} provides the tuning range for the relevant hyperparameters. 
Grid-search is used to tune the hyperparameters of CNN models. The splitting ratios (train-validation-test) are considered as $60\%$-$30\%$-$10\%$ and $60\%$-$20\%$-$20\%$ for the Alaska and Oregon bridges. 

\begin{table}[!h]
\vspace{15pt}
\caption{Top-$5$ CNN models (scour prediction) on validation sets derived by meanMAE policy.}
\resizebox{136mm}{!}{
\begin{tabular}{@{}|c|c|l|c|l|c|l|@{}}
\hline
Bridge & Vanilla Convolutional  & avg. & Dilated Causal& avg. & Fully Convolutional & avg. \\
 & Netwrok (VCN) & MAE & Network (DCN) & MAE & Network (FCN) & MAE \\ \hline
\multirow{5}{*}{230} & vcn-5-64-9-128-0 & 0.090 & dcn-9-64-9-128-0 & 0.107 & fcn-5-64-7-64-0 & 0.086 \\
 & vcn-7-64-9-128-0 & 0.091 & dcn-5-64-5-128-0 & 0.107 & fcn-3-64-5-128-0 & 0.086 \\
 & vcn-7-128-9-256-0 & 0.092 & dcn-7-64-7-128-0 & 0.108 & fcn-9-64-7-128-0 & 0.086 \\
 & vcn-7-128-9-256-0.2 & 0.093 & dcn-3-64-3-64-0 & 0.109 & fcn-3-128-5-128-0 & 0.087 \\
 & vcn-7-64-9-128-0 & 0.093 & dcn-3-128-5-64-0 & 0.109 & fcn-5-64-7-128-0 & 0.088 \\
 \hline
\multirow{5}{*}{539} & vcn-9-128-9-64-0.2 & 0.382 & dcn-5-128-5-64-0 & 0.323 & fcn-3-256-5-128-0 & 0.315 \\
 & vcn-9-128-7-64-0 & 0.384 & dcn-5-64-3-64-0 & 0.327 & fcn-7-256-7-128-0.2 & 0.317 \\
 & vcn-9-64-9-64-0.2 & 0.389 & dcn-7-128-5-64-0.2 & 0.328 & fcn-5-128-7-256-0 & 0.318 \\
 & vcn-9-64-9-64-0 & 0.389 & dcn-5-256-7-256-0 & 0.329 & fcn-9-64-7-128-0 & 0.319 \\
 & vcn-9-64-7-64-0 & 0.391 & dcn-7-64-7-128-0 & 0.330 & fcn-9-128-7-64-0.2 & 0.319\\
 \hline
 \multirow{5}{*}{Luckiamute} & vcn-3-64-3-64-0.2 & 0.525 & dcn-5-256-3-256-0.2 & 0.305 & fcn-5-128-5-256-0.0  & 0.278 \\
 & vcn-5-128-5-64-0.0 & 0.549 & dcn-3-128-5-256-0.2 & 0.308 & fcn-5-256-5-256-0.0  & 0.282 \\
 & vcn-5-64-5-64-0.2 & 0.583 & dcn-5-256-5-256-0.2 & 0.309 & fcn-5-256-5-256-0.2  & 0.283 \\
 & vcn-5-64-7-128-0.2 & 0.606 & dcn-3-256-3-256-0.2 & 0.309 & fcn-3-128-5-256-0.0  & 0.285 \\
 & vcn-5-64-7-64-0.2 & 0.610 & dcn-5-128-5-256-0.0 & 0.310 & fcn-5-256-3-256-0.0  & 0.286\\
 \hline
\end{tabular}
}
\label{table:CNN_Results}
\end{table}

Table \ref{table:CNN_Results} summarizes the performance of the CNN models for scour prediction (Sonar). The FCN variant yields the best performances, as evidenced by the lower MAE scores in Table \ref{table:CNN_Results}, in comparison with the VCN and DCN models for both Alaska and Oregon bridges. The best-performing CNN models achieve competitive accuracy for both Alaska and Oregon bridges, compared with LSTM. 

\begin{table}[!htb]
\centering
\begin{tabular}{c}
      \resizebox{\columnwidth}{!}{\includegraphics{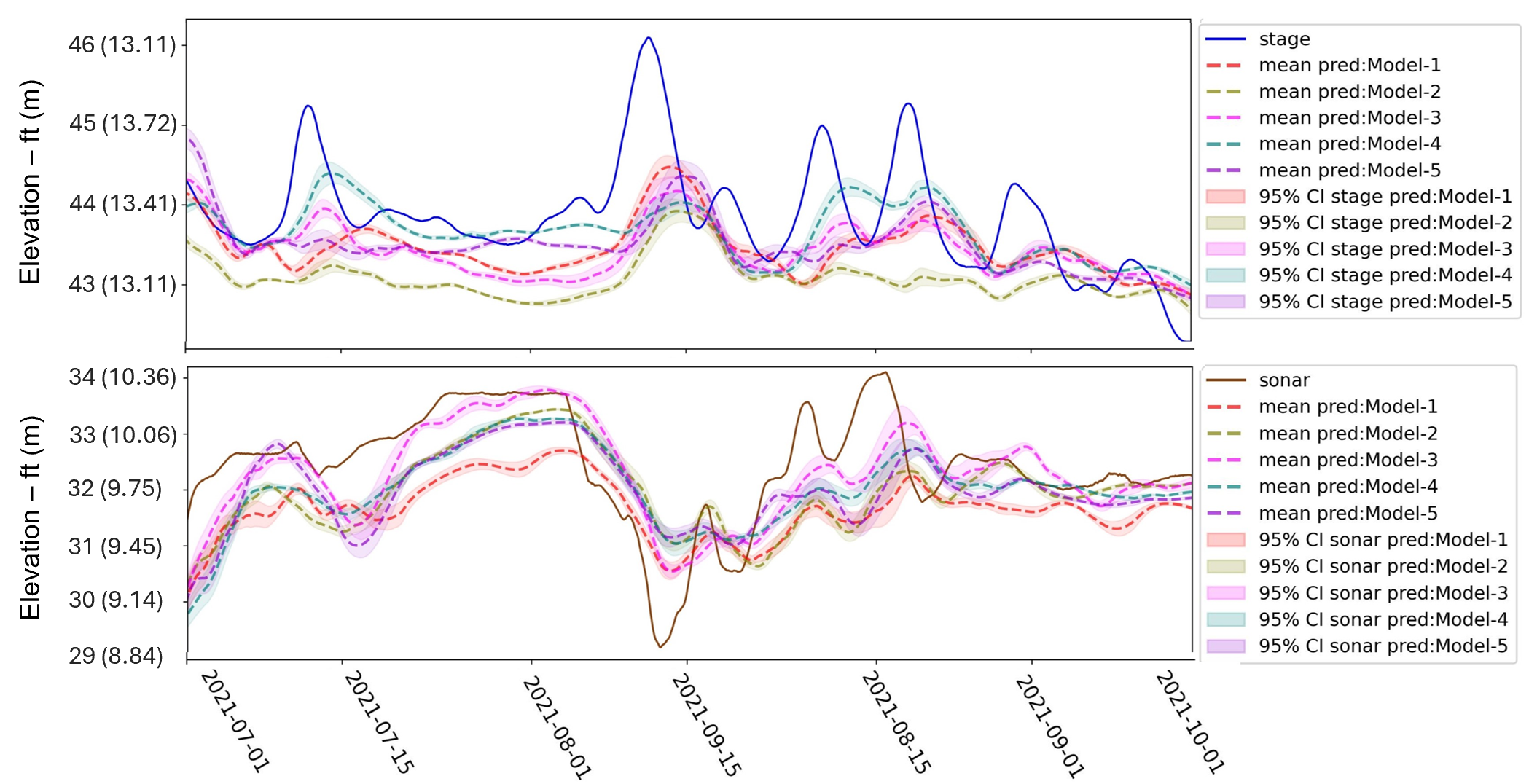}}
\end{tabular}
\captionof{figure}{Stage (top) and Sonar (bottom) predictions of top-$5$ FCN models selected by the meanMAE policy, mean and 95\% confidence intervals over the test dataset - Alaska $230$ bridge.}
\label{Fig:FCN_prediction_230}
\end{table}

\begin{table}[!htb]
\centering
\begin{tabular}{c}
      \resizebox{\columnwidth}{!}{\includegraphics{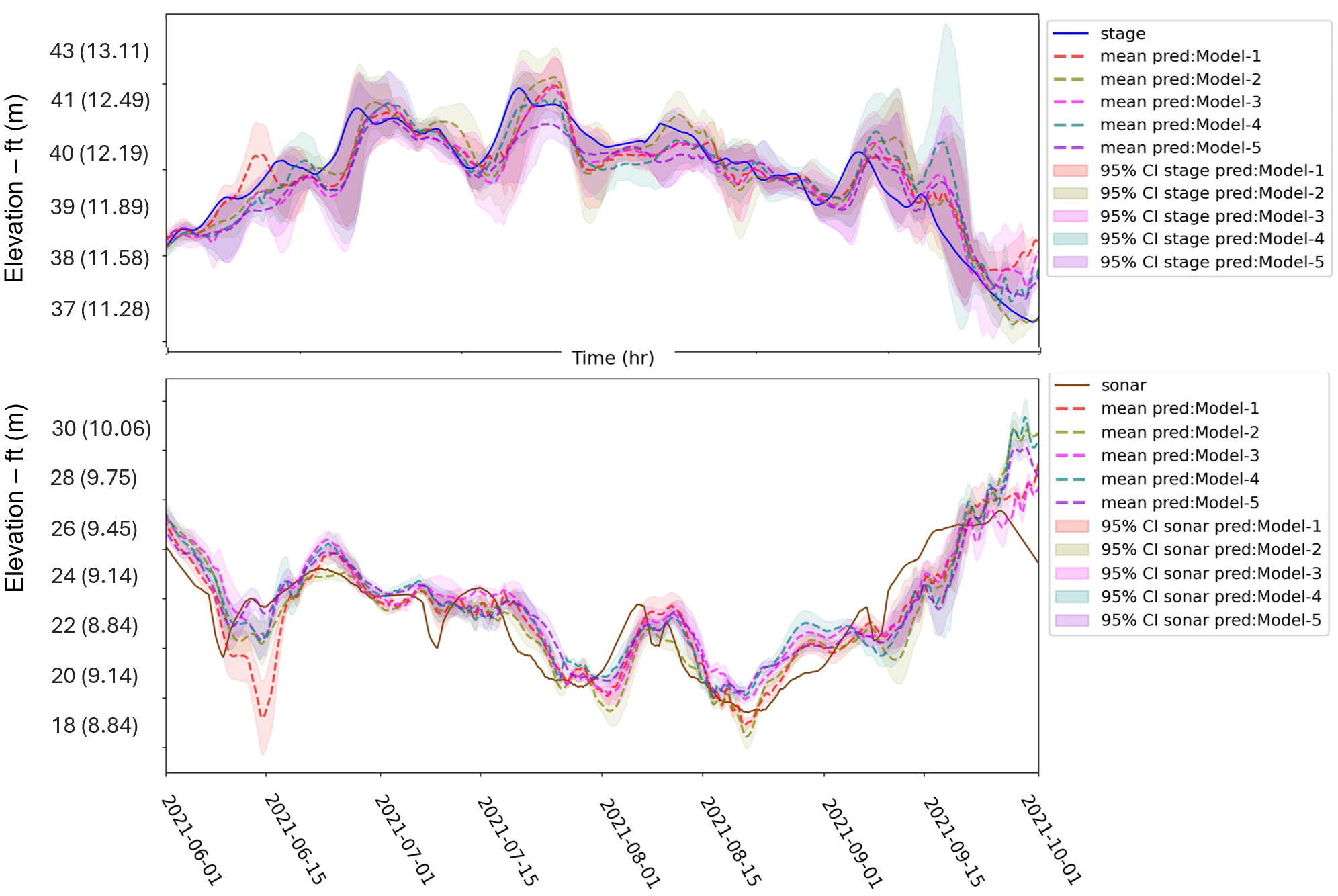}}
\end{tabular}
\captionof{figure}{Stage (top) and Sonar (bottom) predictions of top-$5$ FCN models selected by the meanMAE policy, mean and 95\% confidence intervals over the test dataset - Alaska $539$ bridge.}
\label{Fig:FCN_prediction_539}
\vspace{5pt}
\end{table}

\subsubsection{Scour Forecast for Alaska Bridges}\label{sec:CNN_Results_Alaska}

Fig. ~\ref{Fig:FCN_prediction_230} and Fig. ~\ref{Fig:FCN_prediction_539} demonstrate the time-series prediction of Sonar and Stage by the top-5 FCN models across the test dataset. The CNN models demonstrate a reasonable performance, comparable with LSTM models, capturing the trend of scour and filing a week in advance.

The CNN model with full convolution from Equation~\ref{eq:conv_padded}
shows that it outperforms other convolutional strategies in generalization and captures both Sonar and Stage trends. Particularly, DCN models fail to capture the scour and filling trends as compared to the VCN and FCN. This implies that skipping the intermediate observations by the dilated convolutions from Equation ~\ref{eq:conv_dilatedcasual} did not result in higher accuracy. One reason behind the superior performance by FCN can be the inclusion of the edging values while making batch-to-batch transitions via the $0$-$padded$ convolution. This ensures no loss of information while learning the underlying scour patterns.


\subsubsection{Scour Forecast for Oregon Results}\label{sec:CNN_Results_Oregon}

Fig. ~\ref{Fig:FCN_prediction_Luckiamute} illustrates the performance of top-5 FCN models over the test dataset, between April 2021 and July 2021. Similar to the LSTM models, the CNN models perform less accurately for Oregon data compared with Alaska, particularly in capturing the scour and fill trends. The error of mean Sonar prediction varies between 1.5 to 2.5ft (0.5 to 0.75m). 

\begin{table}[!htbp]
\centering
\begin{tabular}{c}
      \resizebox{\columnwidth}{!}{\includegraphics{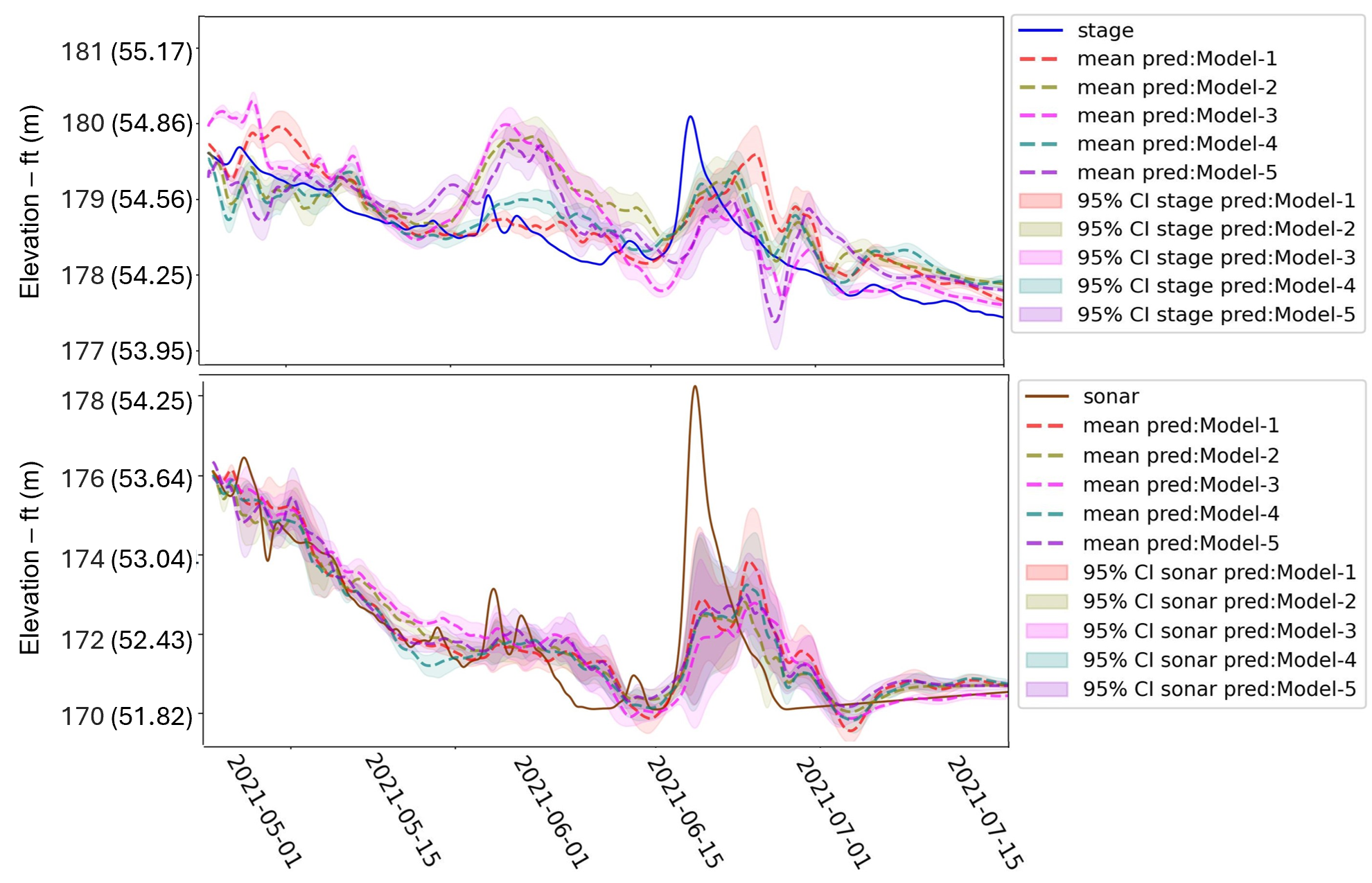}}
\end{tabular}
\captionof{figure}{Stage (top) and Sonar (bottom) predictions of top-$5$ FCN models. The mean and 95\% confidence intervals are considered on the test dataset - Oregon Luckiamute bridge.}
\label{Fig:FCN_prediction_Luckiamute}
\end{table}


\subsection{Feature Combination Analysis (Experiment 4) }\label{sec:wrapper}
\begin{figure}[!htb]
    \centering
    \begin{subfigure}{0.35\textwidth}
        \includegraphics[width=\textwidth]{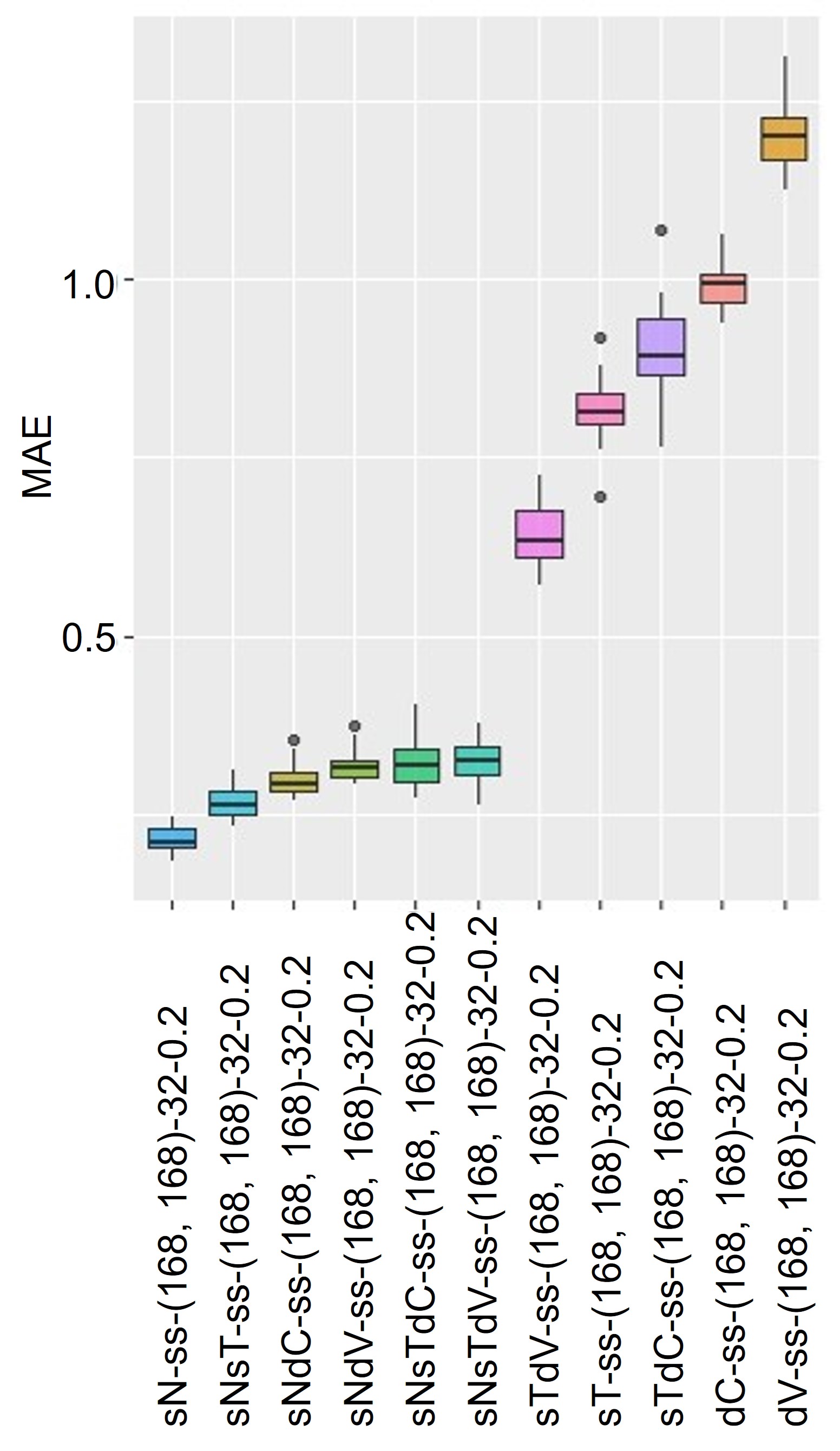}
        \caption{Alaska 539}
    \end{subfigure}
    \begin{subfigure}{0.35\textwidth}
        \includegraphics[width=\textwidth]{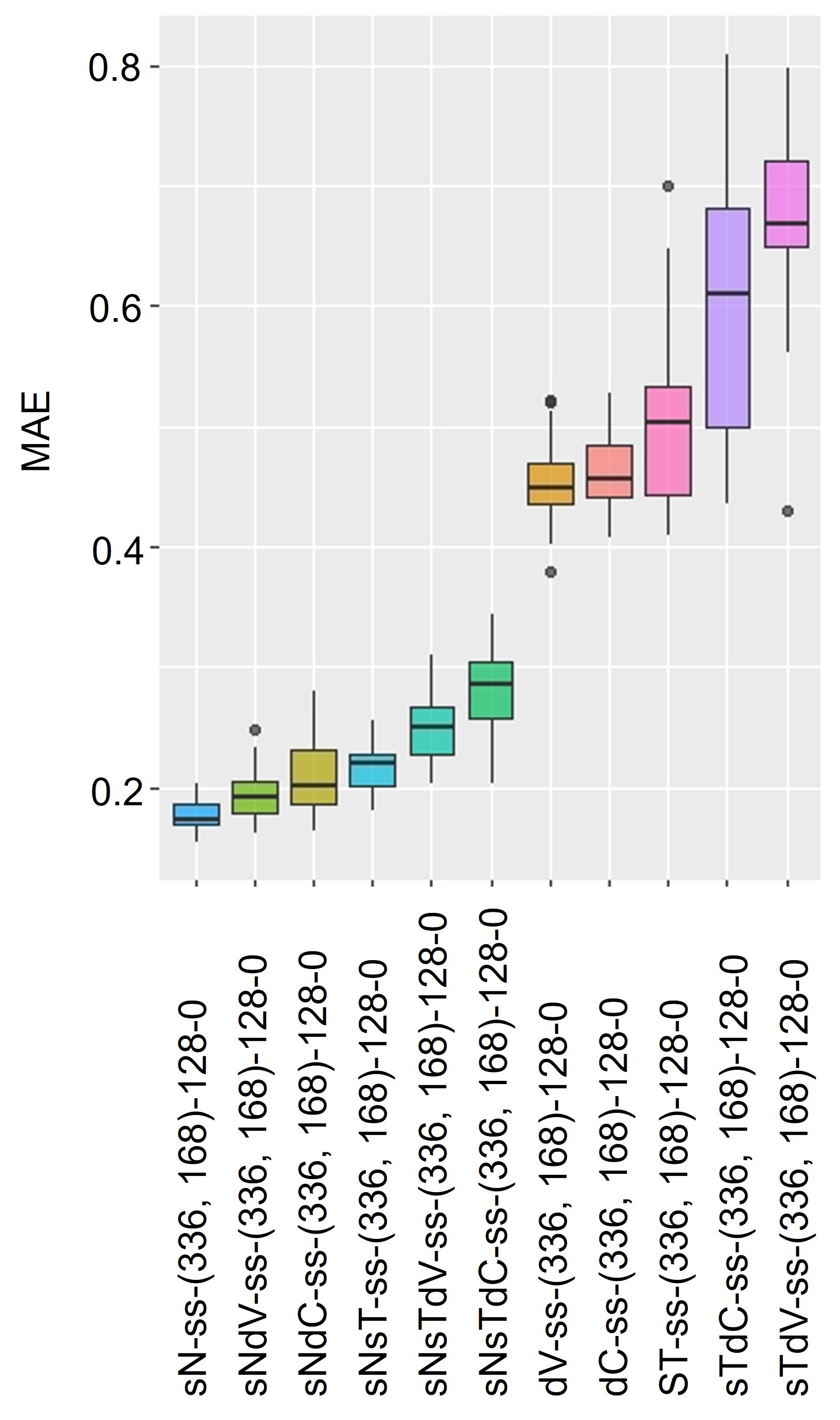}
        \caption{Alaska 230}
    \end{subfigure}
\vspace{1cm}
\caption{Feature impact evaluation - Alaska. The first term in model configurations indicates the feature combination.}
\vspace{10pt}
\label{fig:Alaska_ftimpact}
\end{figure}

\begin{figure}[!htb]
    \centering
    \begin{subfigure}{0.3\textwidth}
        \includegraphics[width=\textwidth]{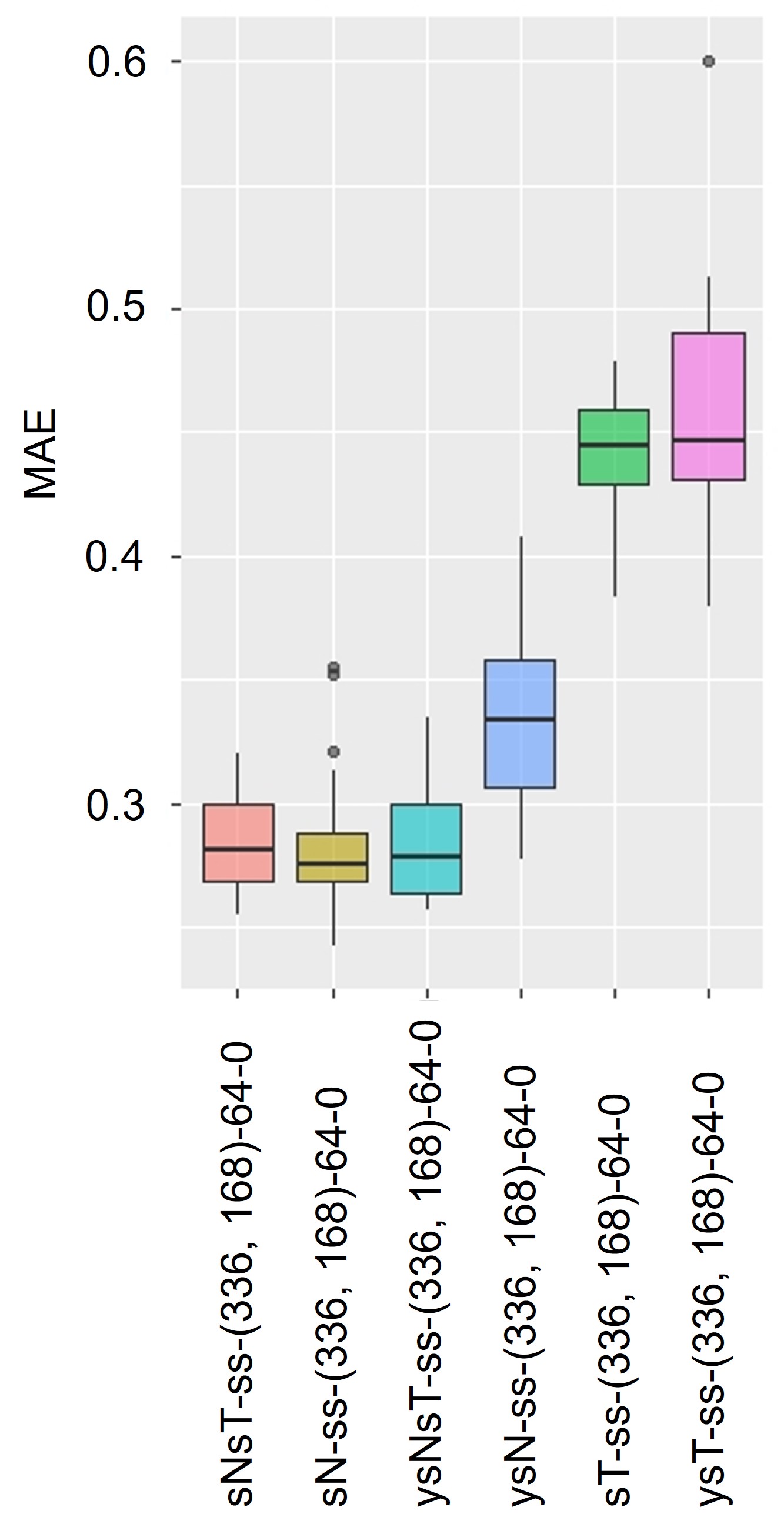}
        \caption{Oregon Luckiamute}
    \end{subfigure}
    \begin{subfigure}{0.3\textwidth}
        \includegraphics[width=\textwidth]{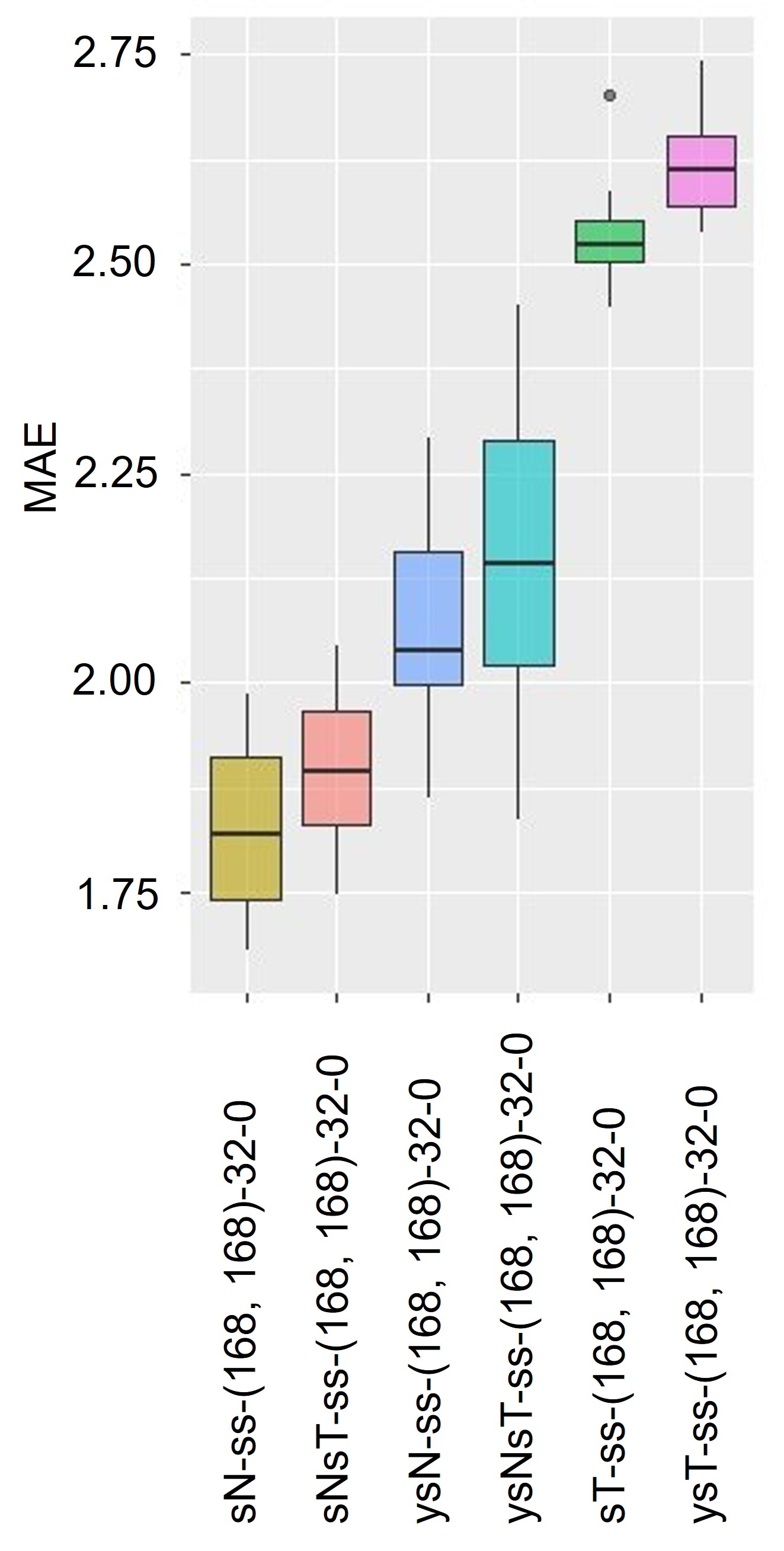}
        \caption{Oregon Trask}
    \end{subfigure}
\vspace{1cm}
\caption{Feature impact evaluation - Oregon. The first term in model configurations indicates the feature combination.}
\label{fig:Oregon_ftimpact}
\end{figure}

Different combinations of features are considered as input to LSTM models to evaluate their impact in scour prediction (wrapper method). The best Single-Shot LSTM configurations were trained for this purpose over two Alaska and Oregon bridges as shown in Figs. \ref{fig:Alaska_ftimpact} and \ref{fig:Oregon_ftimpact}. Note that for Oregon bridges, the discharge data was not; therefore, \textit{dC} \textit{dV} are not included in the impact analysis.

The first term in model configuration nomenclatures indicates the feature combination, e.g. \textit{sNsTdV}, indicates \textit{\{Sonar, Stage, Discharge\}} input feature combination (refer to Section \ref{sec:feat_engineering} for features notations).

These plots show that the input feature sets containing $sN$ (Sonar) consistently outperform the feature sets without $sN$ for all bridges. The models with $sN$ as the only input feature shows the best performance for all bridges. The groups $sNdV$ and $sNdV$ show the second smallest error for all bridges, showing comparable performance as $sN$ alone. As expected, Discharge and eVelocity have an equivalent impact on prediction performance as $sNdC$ and $sNdV$ models show consistently similar error ranges. Also, Stage seems to have a similar impact on prediction as Discharge and eVelocity, as observed in very close error ranges among $sNsT$ and $sNdC$, and $sNdV$. Therefore, adding velocity features to DL models in Alaska does not seem to improve significantly scour forecasting as initially expected. This further fortifies the results reported in ~\cite{phase2} about stage time series having a high correlation with discharge time series, therefore embodying velocity characteristics required for LSTMs to learn the scour patterns.

For the Oregon bridges, the scour models do not obtain higher accuracy by including the time features denoted as \textit{y}, similar to Alaska case studies in ~\cite{phase2}. Input feature sets: $sN$ and $sNsT$ obtain similar accuracy in predicting scour. Similar to Alaska bridges, LSTM models show poor performance when the input feature set does not contain the Sonar values.

The strong dependency among the recent and past values of the sonar readings is evidenced by the lower MAE of the scour model with Sonar as the only feature and the high autocorrelation coefficient of sonar readings. However, Stage is shown to be more impactful for scour prediction in the case of the Luckimaute bridge. This can be related to a lack of consistent periodic scour and filing patterns as opposed to Alaska bridges and Trask bridges with tidal river bed fluctuations. Nevertheless, it is evident from this experiment that flow-related times series on their own, i.e., stage, discharge and velocity, are inadequate for reliable scour prediction. 

\section{Conclusions}\label{conclusions}
This paper investigated prominent Deep Learning (DL) algorithms for scour forecasting based on historic sensor monitoring data, including three variants of LSTM and CNN, for case-study bridges in Alaska and Oregon. The primary focus was to address the challenges of DL models for real-time scour prediction, including 1) performance optimization of algorithms and hyperparameter tuning, 2) generalisation to diverse locations with distinct geological/geomorphological and hydraulic/flow characteristics, and 3) impact of various combinations of sensor features relevant to scour.

The top-performing LSTM models achieved mean absolute error (MAE) ranging from 0.1m (in Alaska) to 0.5m (in Oregon). The best CNN model, the FCN variant, showed similar error ranges (0.1m to 0.75m). Both DL algorithms showed promising performance in forecasting scour for bridges in Alaska, with CNN models, particularly the FCN variant, achieving slightly lower prediction errors. However, CNN was trained with significantly lower computational costs running on the same GPU clusters.  Higher errors were observed for Oregon bridges compared to Alaska, especially in capturing peaks of scour and filling episodes. This was related to the fundamentally different scour process in Oregon bridges, such as tidal flows and the presence of coastal bed forms, and the lack of substantial historical data. In addition, Oregon bridges did not show consistent periodical and seasonal flow and scour patterns as observed in Alaskan bridges. 

To improve the accuracy of DL models for Oregon, we implemented a sequential training approach integrated with transfer learning. This training approach proved effective in improving the performance, while reducing the computational time.

For LSTM models, the single-shot model consistently outperformed the two-layer single-shot and feedback variants. Regarding the CNN, the FCN achieved lower MAE values on all bridges compared to the Vanilla Convolutional Network and Dilated Causal Network. Additionally, our feature impact analysis using the wrapper method showed adding Discharge, and Velocity time series to Stage and Sonar did not improve the performance of DL models in Alaska. An interesting observation consistent across all bridges in Oregon and Alaska was that the models without Sonar (bed elevation) time series showed significantly higher errors. In fact, models with Sonar as the only input feature obtained similar (and in some cases slightly better) scour prediction accuracy compared with benchmark models with the Sonar plus Stage feature set. For Oregon bridges, the time features did not improve the performance of DL models similar to what was observed previously for Alaska.

The random-search method with mean MAE and Bagging heuristics introduced in this study proved to be efficient alternatives to grid-search for hyperparameter tunning. The top LSTM configurations identified by the random-search with higher sample sizes (\%67) resulted in finding the most optimal configurations, while saving significantly on computational cost.

In conclusion, this study highlights the potential of deep learning models, particularly LSTM and CNN variants, for accurate scour forecasting in diverse bridge locations. However, challenges remain in adapting these models to fundamentally different scour processes and limited historical data, as observed in the Oregon bridges. Future research should focus on developing more robust and adaptable models that can handle varying scour characteristics and data limitations. Additionally, integrating domain knowledge and physics-based modelling approaches with data-driven techniques could further improve the interpretability and generalization of scour prediction models. Finally, the exploration of advanced hyperparameter optimization techniques and transfer learning strategies may help reduce computational costs and improve the efficiency of model development for scour forecasting applications.

\section{Acknowledgement}\label{acknowledgement}
The funding for this research was provided by The University of Melbourne and Arup Global Research. We would like to express our gratitude to Greg Lind from USGS and Kira Glover-Cutter from Oregon DOT for providing the data and insights throughout this research. We also extend our thanks to Dr Bo Wang for his help in preparing this manuscript.


\bibliography{ScourPrediction}
\end{document}